%% file: main.tex
\documentclass[11pt, a4paper, twocolumn, confidential, copyright, goog]{google}

\usepackage{amssymb}
\usepackage[authoryear, sort&compress, round]{natbib}
\usepackage{array}    
\usepackage{multirow}      
\usepackage{booktabs}      
\usepackage{tabularx}
\usepackage{ragged2e}

\newcolumntype{L}[1]{>{\raggedright\arraybackslash}p{#1}} 
\newcolumntype{C}[1]{>{\centering\arraybackslash}m{#1}}  
\newcolumntype{R}[1]{>{\raggedleft\arraybackslash}m{#1}} 
\newcolumntype{J}[1]{>{\justifying\arraybackslash}p{#1}}  

\let\copyrightext\relax
\let\today\relax

\title{Large Language Models in Argument Mining: A Survey}

\correspondingauthor{hao.li-2@manchester.ac.uk}

\author[1]{Hao Li}
\author[1,2]{Viktor Schlegel}
\author[1]{Yizheng Sun}
\author[1]{Riza Batista-Navarro}
\author[1]{Goran Nenadic}

\affil[1]{Department of Computer Science, University of Manchester, UK} 
\affil[2]{Imperial Global Singapore, Imperial College London, Singapore}

\begin{abstract}

\noindent\hfill \textit{\textcolor{blue!50!black}{This survey covers literature up to September 27, 2025.}} \hfill\mbox{}

\vspace{1em}  

Large Language Models (LLMs) have fundamentally reshaped Argument Mining (AM), shifting it from a pipeline of supervised, task-specific classifiers to a spectrum of prompt-driven, retrieval-augmented, and reasoning-oriented paradigms. Yet existing surveys largely predate this transition, leaving unclear how LLMs alter task formulations, dataset design, evaluation methodology, and the theoretical foundations of computational argumentation.
In this survey, we synthesise research and provide the first unified account of AM in the LLM era. We revisit canonical AM subtasks, i.e. claim and evidence detection, relation prediction, stance classification, argument quality assessment, and argumentative summarisation—and show how prompting, chain-of-thought reasoning, and in-context learning blur traditional task boundaries. We catalogue the rapid evolution of resources, including integrated multi-layer corpora and LLM-assisted annotation pipelines that introduce new opportunities as well as risks of bias and evaluation circularity. Building on this mapping, we identify emerging architectural patterns across LLM-based AM systems, and consolidate evaluation practices spanning component-level accuracy, soft-label quality assessment, and LLM-judge reliability. Finally, we outline persistent challenges, including long-context reasoning, multimodal and multilingual robustness, interpretability, and cost-efficient deployment, and propose a forward-looking research agenda for LLM-driven computational argumentation.

\end{abstract}

\begin{document}

\maketitle

\input{section/1_intro}

\input{section/2_methodology}
\input{section/4_dataset}
\input{section/3_task}
\input{section/5_evaluation}

\input{section/6_methods}
\input{section/7_future}
\input{section/8_conclusion}

\bibliography{main}

\end{document}

%% file: section/1_intro.tex
\section{Introduction}
 Argument Mining (AM) seeks to automatically identify, structure, and interpret argumentative discourse by extracting claims, premises, and the inferential relations that connect them \citep{patel2024machine}. As a research area situated at the intersection of natural language processing (NLP), discourse analysis, and computational argumentation, AM underpins applications ranging from online content moderation and fact verification to legal analytics, scientific peer review, and writing-support systems\cite{DBLP:journals/coling/LawrenceR19}. Classical AM methodologies have largely adopted task-specific supervised pipelines, wherein component detection, relation classification, and argument-quality assessment are treated as independent modelling problems supported by carefully designed annotation schemes and modestly sized corpora\citep{DBLP:journals/coling/LawrenceR19}.

Despite its clear value, manual argument annotation remains labor-intensive and inherently unscalable: crafting detailed argument maps can consume hundreds or even thousands of hours for relatively short texts. To cope with the volume of data generated online, research in Argument Mining (AM) has accelerated significantly since around 2013, marked by the launch of dedicated workshops at ACL and themed tutorials at top-tier conferences including IJCAI, ACL, and ESSLLI \citep{DBLP:conf/ijcai/CabrioV18}. Foundational surveys—most notably those by \cite{DBLP:journals/coling/LawrenceR19}, which comprehensively cataloged AM techniques, datasets, and challenges \citep{DBLP:journals/coling/LawrenceR19}; Patel \cite{DBLP:journals/ijcini/PeldszusS13}, who systematically bridged classical argument-diagramming methods with machine-readable annotation pipelines \citep{DBLP:journals/ijcini/PeldszusS13}; and \citet{DBLP:journals/ipm/LytosLSB19}, who documented AM’s adaptation to less formal, user-generated text \citep{DBLP:journals/ipm/LytosLSB19}—have provided essential context for early developments.

\begin{figure*}[t]
    \centering
    \includegraphics[width=\linewidth]{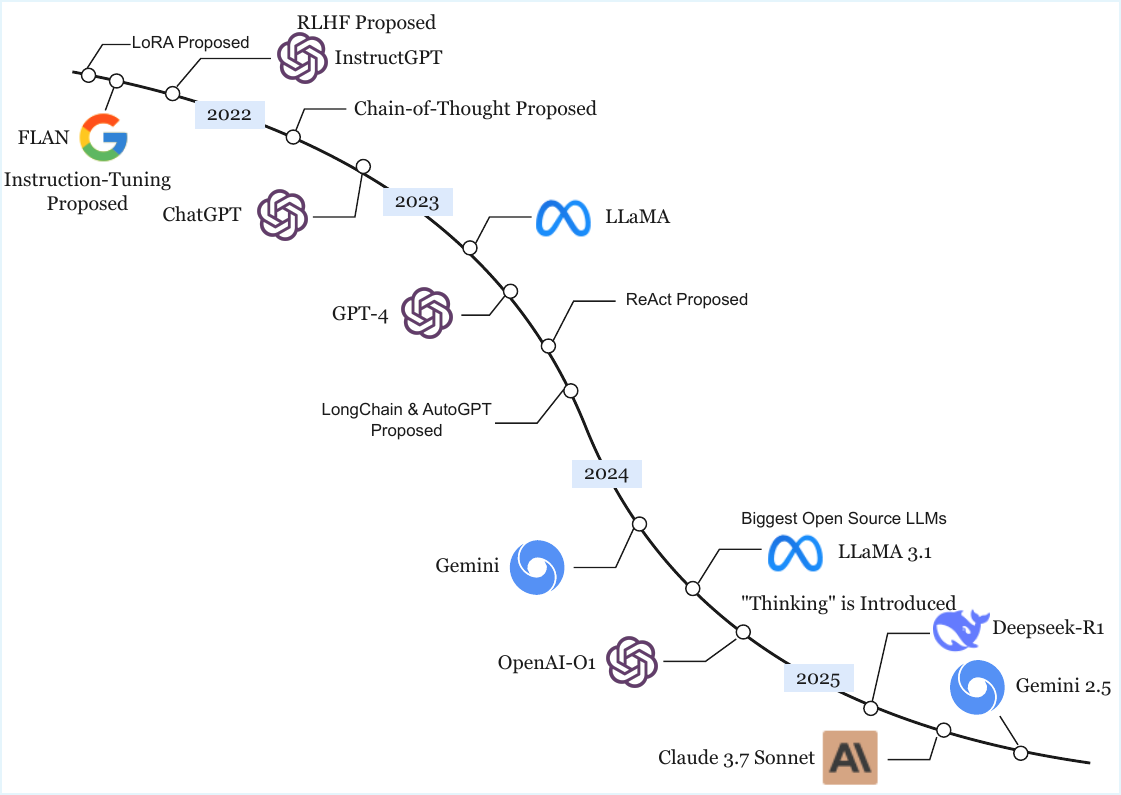}
    \caption{Key large language models and techniques influencing argument mining.}
    \label{fig:llm_timeline}
\end{figure*}

Building on these foundations, subsequent studies have enriched the field with domain- and method-specific insights. \citet{DBLP:conf/ijcai/CabrioV18} offered a data-driven reflection on the field’s early corpus and model evolution , while targeted surveys explored annotation practices, tool ecosystems, and argument quality assessment. For example, \citet{patel2024machine} traces the integration of traditional machine learning with emerging neural techniques; \citet{DBLP:journals/it/SchaeferS21} focus on Twitter; \citet{DBLP:conf/naacl/HuaNBW19} and \citet{DBLP:conf/aaai/0001FBBQZDSM021} examine peer-review text; and \citet{guida2025llms} assess the performance of LLMs on enforcing topic-specific argument tasks in online comments \citep{DBLP:journals/corr/abs-2503-00847, DBLP:conf/acl/ChenCLB24}.

However, despite this breadth, most existing reviews remain anchored in the pre‑LLM era. They emphasize supervised learning and feature engineering, formal annotation schemas, or segmented task pipelines, and so far have only cursorily addressed the revolutionary role of pre-trained language models. In response, a new wave of empirical literature has emerged, exemplified by work on LLM-driven argument summarization, argument extraction, and relationship classification \citep{DBLP:conf/acl/ChenCLB24, DBLP:journals/corr/abs-2503-00847, guida2025llms}. Yet, there is still no review that systematically integrates these advances (shown in Figure \ref{fig:llm_timeline}), examines how LLMs reshape the boundaries and synergies of AM subtasks, examining how advances in one area reshape assumptions in another, and reconnects these technical developments to foundational argumentation theory and annotation practice. We treat argument mining as an \emph{interlocking system} whose components—tasks, data, metrics, and models evolve together under the pressure of foundation-model capabilities. 

This survey aims to fill this gap by reframing argument mining (AM) in the era of foundation models. Rather than treating AM tasks as isolated modules, we conceptualize the field as an interdependent system in which tasks, evaluation methodologies, and data resources co-evolve under the influence of large language models (LLMs). Our contributions are threefold. \emph{(i)} \textbf{Dataset landscape consolidation.}
Chapter \ref{sec:dataset} provides a unified overview of datasets across argumentative discourse types, languages, and modalities. We highlight how LLM-assisted annotation pipelines, synthetic data generation, and multi-source evidence corpora expand coverage while introducing new questions regarding reliability, bias, and inter-annotator dynamics. \emph{(ii)} \textbf{Reconceptualising tasks and evaluation.}
Chapter \ref{sec:taxonomy} and and \ref{sec:evaluation} revisits core AM tasks—claim and evidence detection, stance identification, relation parsing, argument quality assessment, and summarisation—showing how prompting, chain-of-thought reasoning, and retrieval-augmented generation increasingly blur traditional task boundaries. Crucially, we pair each task with its modern evaluation paradigm, illustrating how dataset-specific constraints, soft-label metrics, and LLM-based judges reshape what counts as model “performance’’ in practice.
\emph{(iii)} \textbf{Technology survey and agenda-setting.}
Chapter \ref{sec:models} synthesises architectural patterns that characterise LLM-era AM, including instruction tuning, debate-structured prompting, role-conditioned cascades, and graph-augmented decoding. Finally, Chapter \ref{sec:future} outlines key research frontiers: long-context argumentative reasoning, multimodal and multilingual generalisation, bias auditing for synthetic or LLM-generated data, and cost-efficient deployment for real-world argumentative systems.

%% file: section/2_methodology.tex
\section{Methodology}
\label{sec:method}

\begin{figure}[t]
  \centering
  \includegraphics[width=\linewidth]{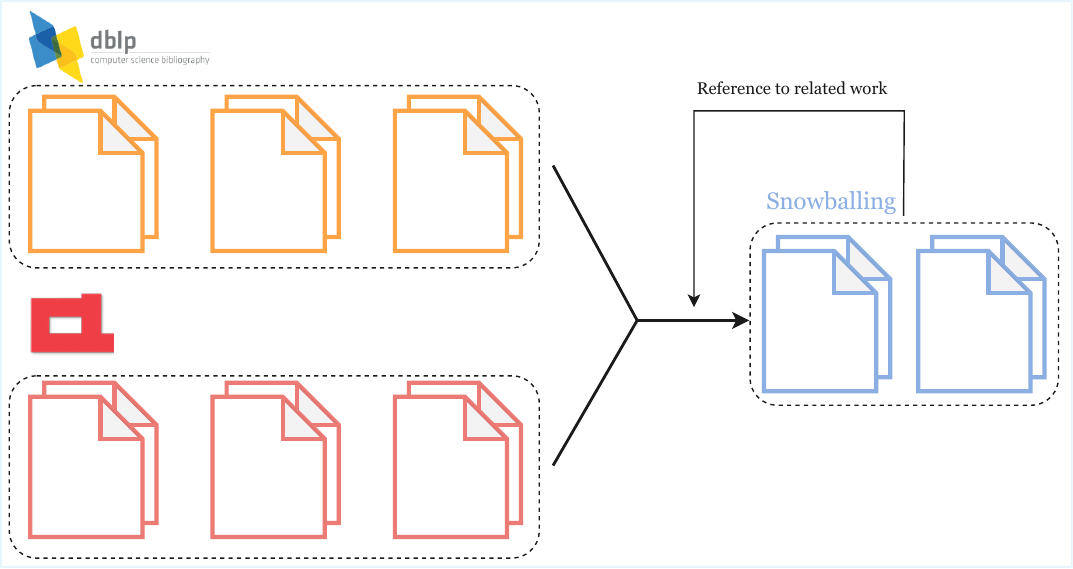}
  \caption{Paper-collection workflow and resulting dataset.}
  \label{fig:collection_pipeline}
\end{figure}

To obtain a comprehensive, up-to-date picture of research on \textit{argument mining} (AM) in the large-language-model era, we followed a three-step protocol inspired by prior survey work \citep{DBLP:conf/emnlp/IvanovaHN24}.  Figure~\ref{fig:collection_pipeline} gives an overview.

\textbf{Sources.} We first manually queried two bibliographic repositories that jointly cover virtually all peer-review venues in computational linguistics and artificial-intelligence research: (i) the Digital Bibliography and Library Project\footnote{\url{https://dblp.org}} (DBLP) and (ii) the ACL Anthology\footnote{\url{https://aclanthology.org}}. Because the terminology used by the field is heterogeneous, we constructed a Boolean query that targets both generic and fine-grained AM vocabulary (e.g. argument, claim, premise, evidence, key point) and action like mining, extraction, detection. Following recent survey conventions, we limited the search to the last ten calendar years (2015--2025, inclusive), specific for recent 4 years of LLMs era. We adopt this ten-year window because several influential techniques—such as LoRA and other parameter-efficient methods—originated earlier but have since become foundational in modern LLM-based argument mining systems. We find around 210 unique publications that match the query and fall into the 10-year window.  We de-duplicated pre-prints against their camera-ready versions by DOI.

\paragraph{Dataset} Many AM papers introduce new corpora or benchmarks that subsequently shape the evaluation landscape.  
We therefore flagged all items whose \textsc{title} or \textsc{booktitle} fields contained any of the strings \textit{dataset}, \textit{corpus}, \textit{benchmark}, \textit{collection}, or \textit{shared task}.  
This yielded around \textbf{70} \emph{dataset papers}, i.\,e., works that either (a) release an original annotated resource or (b) substantially extend an existing one. In this work we focus on \textbf{40} of represent work to discuss.

\paragraph{Iterative Snowballing}
To mitigate false negatives, especially papers that use idiosyncratic task names (e.g.\ ``key-point analysis'', ``argument quality ranking'') we employed one rounds of backward and forward snowballing on the seed set’s references.  The procedure added 38 further works (6.9 \% of the corpus), all of which passed the same year and scope filters.  Because every snowballed item also appears in DBLP or the ACL Anthology, the final corpus size is \textbf{250} research papers.

%% file: section/4_dataset.tex
\section{Dataset and Benchmark}
\label{sec:dataset}

\input{table/dataset_statistics_2020}
\input{table/dataset_statistics_2025}

Annotated corpora have always served a dual function in argument–mining research: they are the \emph{evidence} on which empirical claims are validated, and they are the \emph{scaffolding} that decides which problems can be modelled at all.  
Historically, dataset design followed a “task-first’’ logic—each new sub-task (claim detection, convincingness ranking, sufficiency checking, key-point analysis) first appeared as a corpus release and only thereafter as a stream of modelling papers. Compared with the small, task-specific, English-only corpora that served feature-based models, today’s “LLM-era’’ datasets are larger, multi-label, cross-domain, and often partly machine-generated—properties chosen explicitly to match the breadth, multitask capacity, and data appetite of modern instruction-tuned language models. In this work, we conduct a comprehensive survey of how datasets and benchmarks meet the requirements and adapt to evolution of the LLM era. 

Specifically, we outline it in three successive research phases: (i) \textbf{from task-specific, human‐curated collections to integrated, LLM-centric resource} (Section \ref{ssec:classic_to_integrated}), (ii) \textbf{Using LLM as annotator and generator} (Section \ref{ssec:llm_assisted}, and (iii) an emerging focus on \textbf{scale, modality, and language breadth} (Section \ref{ssec:scale_modality_language}).  We review each phase and close look at over 40 datasets (shown in Table \ref{tab:datasets_statistics_2021} and \ref{tab:datasets_statistics_2025} to figure out the trends and open challenges.

\subsection{From One Task–One Dataset to Integrated Corpora}
\label{ssec:classic_to_integrated}

The first generation of argument-mining resources was intentionally \emph{mono-task}.  
Each corpus addressed a single research question and encoded only the labels required to answer it.  
For instance, UKP \textsc{OpposingArgumentsInEssays} focussed exclusively on segmenting \emph{major claims} and \emph{premises} in student essays and reached expert agreement of $\kappa=.81$ by concentrating annotator effort on two labels alone \citep{DBLP:conf/acl/StabG16}.  
UKP \textsc{ConvArg} framed argument quality as pairwise \emph{convincingness} comparisons, enabling early feature-based rankers and later RoBERTa baselines, but left logical sufficiency and rhetorical style unlabelled \citep{DBLP:conf/acl/HabernalG16}.  
IBM \textsc{InsufficientArguments} captured binary \emph{sufficiency} in 1 000 essay sentences; the narrow scope made it ideal for entailment studies, yet its labels could not generalise to social-media debates or legal opinions \citep{DBLP:conf/eacl/GurevychS17}.

Single-purpose design proved invaluable for \textit{methodological clarity}.  
Because each dataset held only one variable constant, researchers could dissect feature contributions and model architectures with minimal confounds.  
Notably, margin-ranking SVMs on \textsc{ConvArg} delivered a 19-point Pairwise Accuracy gain over bag-of-words in large part because the dataset’s simplicity isolated ranking logic from component detection errors \citep{DBLP:conf/acl/HabernalG16}.  
However, this fragmentation soon impeded cumulative progress.  
Cross-task transfer became almost impossible: a model trained to find claims in UKP essays failed on biomedical abstracts, not because the notion of a claim differed but because no shared label inventory or evaluation split existed across corpora \citep{DBLP:conf/ecai/0002CV20}.  
Moreover, meta-analysis was thwarted by incompatible annotation units (sentence vs. clause) and divergent definitions of “support’’ and “attack’’ \citep{DBLP:conf/acl/ReimersSBDSG19}.

Faced with these limitations—and armed with instruction-tuned large language models capable of multi-task learning—dataset designers began to \emph{integrate} multiple label layers within a single corpus.  
The turning point was \textbf{IAM} (69 k Reddit threads), which simultaneously marks clause boundaries, supporting evidence links, stance polarity, and topic metadata.  
Multi-task RoBERTa trained on IAM improved macro-F\textsubscript{1} by 4–6 points over single-task baselines on both claim detection and stance classification, demonstrating that shared representations benefit from joint supervision \citep{DBLP:conf/acl/ChengBHYZS22}.  
Similarly, \textbf{ARIES} (81 k documents) overlays RST-style discourse graphs on argumentative zones; graph neural networks trained here transfer to the cross-lingual MuLMS-AZ news corpus with only a 5-point F\textsubscript{1} loss, far less than the 15-point drops reported for earlier single-task systems \citep{gemechu2024aries,DBLP:journals/corr/abs-2307-02340}.  
Integrated corpora are not confined to component-level annotation: ORCHID fuses stance, key-point matching, and abstractive summaries in 14 k opinion paragraphs, enabling the first evaluation of summarisation metrics grounded in argumentative structure \citep{DBLP:conf/emnlp/ZhaoWP23}; StoryARG does likewise for narrative argumentation \citep{DBLP:conf/acl/FalkL23}.

The move to multi-facet datasets raises new methodological questions.  
Annotator agreement is now label-specific: in IAM, component spans reach $\alpha=.79$, yet stance falls to .63, and evidence links to .57—numbers rarely reported in earlier corpora.  
Label \emph{interoperability} remains ad hoc: the definition of “premise’’ in IAM (full clause) does not exactly match that in ARIES (minimal span), complicating model reuse.  
Preliminary work on ontology mapping \citep{DBLP:conf/emnlp/IvanovaHN24} shows that automatic label harmonisation lifts zero-shot transfer by 6–8 points, but the field still lacks a canonical schema.

\paragraph{Research gaps and future work.}
(1) \textit{Reliability audits}: integrated corpora should report per-label agreement and analyse cross-label interference (e.g., does stance bias sufficiency?).  
(2) \textit{Shared ontologies}: without a common inventory, the promise of multi-task modelling will stall at dataset borders.  
(3) \textit{Task balancing}: early evidence suggests that over-represented labels dominate learning; dynamic sampling or curriculum learning could mitigate this skew.  
Addressing these issues will turn integrated corpora from mere “data lakes’’ into coherent, theory-driven benchmarks suitable for the next generation of argument-aware language models.

\subsection{LLM-Assisted Creation and Annotation}
\label{ssec:llm_assisted}

Large language models are rapidly becoming collaborators in corpus construction rather than merely downstream consumers.  Their involvement takes three main forms—prompt-driven synthesis, machine-in-the-loop annotation, and fully automatic labelling of difficult variables—and each reshapes long-standing assumptions about cost, bias and reliability.

\paragraph{Prompt-driven synthesis} gained prominence when \citet{DBLP:journals/corr/abs-2304-07666} used GPT-4 to draft thousands of claim–counter-claim pairs across diverse topics.  Although human editors rejected a non-trivial minority for logical incoherence, the retained material exhibits argumentative structures broad enough to boost zero-shot performance on earlier convincingness benchmarks by a noticeable margin—often up to ten percent relative improvement.  Stylistic audits, however, reveal a homogenised, over-polite register and a marked reduction in hedging compared with organically written debate, raising concerns that synthetic data may narrow the stylistic bandwidth of subsequent models \citep{long2024llms}.

\paragraph{Machine-in-the-loop annotation} Humans remain in control but delegate first-pass labelling to an LLM.  The ASE corpus exemplifies this workflow: GPT-4 proposes evidence spans, sufficiency judgements and short summaries, after which human reviewers confirm or amend the suggestions \citep{DBLP:conf/acl/0074WSBMZZWHLN24}.  Verification time falls markedly, yet error analysis shows that chain-of-thought prompting, while improving recall, doubles the rate of subtle hallucinations.  Similar hybrid gains are reported for SurveyKP, where key-point clusters refined by annotators achieve noticeably higher purity than crowd-only baselines at roughly half the labour hours \citep{DBLP:conf/emnlp/EdenKOKSB23}.  The flip side is \emph{automation anchoring}: reviewers tend to accept stylistic quirks or soft politeness biases introduced by the model, an effect already quantified in appropriateness studies by \citet{DBLP:conf/acl/ZiegenbeinSLPW23}.

\paragraph{Automatic Judge} A third trend outsources to automatic judgements LLMs when labels are cognitively heavy or prone to low consensus.  In ConQRet, GPT-4 decides whether retrieved passages “fully answer’’ complex questions before experts handle only borderline cases \citep{DBLP:conf/naacl/DholeSA25}.  Agreement with specialists is high overall, yet topic-sensitive analyses show the model penalises rhetorical questions more harshly than humans—a bias echoing manipulation-check findings on emotional language by \citet{DBLP:journals/corr/abs-2503-00024}.  Moreover, when the same model family later evaluates systems trained on these labels, circularity becomes a real possibility: sufficiency scores judged by GPT-4 appear several points higher when the evaluator is also GPT-4 than when a conventional RoBERTa baseline is used \citep{DBLP:conf/naacl/LiuFC24}.

\paragraph{Research Gaps and Future Works} These developments create new methodological obligations.  Few corpora publish detailed “data cards’’ that document prompt templates, sampling temperatures, rejection rates or post-editing guidelines, leaving provenance opaque and bias assessment difficult.  Systematic audits are needed to understand how demographic, topical and stylistic biases propagate from model to dataset.  Equally pressing is the design of adversarial test suites: models trained solely on clean, LLM-generated text degrade sharply on noisy social-media arguments, suggesting that synthetic and natural data must be blended with care.  Finally, the legal status of machine-generated content remains unsettled; future releases should clarify licensing to safeguard downstream reuse.

In sum, LLM assistance lowers annotation cost and widens the range of available labels, yet it also imports model-specific artefacts and raises questions about bias, provenance and evaluation circularity.  Addressing these issues through transparent documentation, rigorous bias audits and adversarial evaluation will be vital if LLM-centred datasets are to serve as reliable foundations for the next generation of argument-mining models.

\subsection{Scale, Modality, and Language}
\label{ssec:scale_modality_language}

As argument–mining goals have broadened, corpus builders have sought breadth in three orthogonal directions—\emph{size}, \emph{modality}, and \emph{language}.  Each dimension has expanded unevenly, yielding fresh modelling opportunities but also exposing new blind spots.

\paragraph{Size: from thousands to hundreds of thousands.}
The median pre-LLM corpus contained only a few thousand annotated instances, limiting the depth of neural fine-tuning.  Integrated resources now appear at an order‐of-magnitude larger scale.  Debate platform dumps such as \textsc{Kialo} \footnote{https://www.kialo.com/}, with well over a quarter-million crowd-moderated nodes, allow graph-level experiments in argumentation theory that were inconceivable when researchers relied on essay corpora \citep{agarwal2022graphnli}.  Large scale is not merely a matter of quantity: it supports adversarial splitting, more granular error analysis, and realistic low-resource simulations.  Yet volume can disguise annotation drift; agreement figures reported for the first 10 

\paragraph{Modality: beyond plain text.}
Argumentative discourse is rarely confined to prose; tone of voice, slide layouts, or imagery all shape persuasiveness.  Two speech–text corpora IBM’s \textsc{Recorded Debating} and the VivesDebate-Speech set align transcribed turns with audio and meta-data, enabling first forays into prosody-aware argument strength \citep{DBLP:conf/acl/OrbachBTLJAS20,DBLP:conf/emnlp/Ruiz-DolzS23}.  The Image-Argument pilot goes further, annotating visual claims and layout cues in news pages \citep{DBLP:conf/argmining/LiuEZL23}.  Early multimodal baselines show detectable gains when acoustic or layout features are fused with text, but they also reveal new annotation pain-points: irony carried by image–caption incongruence or persuasive pauses in speech remain largely uncaptured.  More importantly, evaluation metrics such as “pairwise convincingness’’ have yet to be re-interpreted for cross-modal signals.

\paragraph{Language: cautious steps toward multilinguality.}
English continues to dominate, but a second tier of languages,German, Dutch, Spanish, French, Italian, and Chinese, has emerged.  Parallel corpora such as \textsc{XArgMining} demonstrate that simple cross-lingual projection recovers much, but not all, of English performance \citep{DBLP:conf/emnlp/Toledo-RonenOBS20}.  Specialised bilingual resources add depth: \textsc{FinArg} pairs English and domain-specific finance jargon with argumentative structure \citep{alhamzeh2022s}, whereas CEAMC offers Chinese–English medical arguments linked by explicit claim–evidence pairs \citep{DBLP:conf/emnlp/RenWLZZYZBL24}.  Nevertheless, most multi-task corpora remain monolingual, so joint modelling of component, stance and quality across scripts is still aspirational.  Preliminary cross-lingual experiments suggest that stance and sufficiency transfer degrade much faster than claim segmentation, hinting that cultural framing affects higher-order argumentative facets more than low-level span boundaries.

\paragraph{Research gaps and future work.}
These expansion axes rarely intersect: no existing corpus is simultaneously large-scale, multimodal, and multilingual. This limits robustness testing—for example, a model trained on English audio debates cannot be evaluated on Chinese medical videos. Future benchmarks should therefore pursue cross-dimensional design, even at reduced scale. Equally urgent is the development of evaluation metrics that remain meaningful when text, audio, and images interact; without such metrics, improvements in one modality may conceal failures in another.

In short, corpora have become larger, richer, and more linguistically diverse, yet development remains patchwork. Addressing the gaps between scale, modality, and language will be essential for argument mining to generalise beyond Anglophone, text-only settings into the multimodal and multilingual environments where real-world argumentation flourishes.

%% file: table/dataset_statistics_2020.tex
\begin{table*}[htbp]
\small\setlength{\tabcolsep}{5pt}

\begin{tabular}{@{}C{2.8cm}C{1cm}C{1cm}C{1cm}C{1cm}C{5cm}C{1.2cm}C{1cm}@{}}
\toprule
\textbf{Dataset} & \textbf{Year} & \textbf{Size} & \textbf{Lang.} &
\textbf{Modality} & \textbf{Main Tasks} & \textbf{LLM Origin\footnotemark} & \textbf{Open Source}\\
\midrule

ArgMinNews & \citeyear{DBLP:conf/emnlp/Eckle-KohlerKG15} & 80 & DE & Text &
Argument Relation Identification & & \\

UKP OpposingArgumentsInEssays & \citeyear{DBLP:conf/acl/StabG16} & 402 & EN & Text &
Claim Detection & & \ding{51} \\

UKP ConvArg & \citeyear{DBLP:conf/acl/HabernalG16} & 16k & EN & Text & Convincingness Checking & & \ding{51}\\

UKP InsufficientArguments &
\citeyear{DBLP:conf/eacl/GurevychS17} & 1 k & EN & Text &
Sufficiency Checking & & \ding{51}\\

IBM Claim Stance Dataset & \citeyear{DBLP:conf/eacl/SlonimBSBD17} & 2.3k & EN & Text & Stance Detection & & \ding{51} \\

IBM Claim Sentences Search & \citeyear{DBLP:conf/coling/LevyBGAS18} & 1.49M & EN & Text &
Claim Detection & & \ding{51} \\

IBM Evidence Quality & \citeyear{DBLP:conf/acl/GleizeSCDMAS19} & 5.6K & EN & Text & Claim Detection & & \ding{51}  \\

PERSPECTRUM & \citeyear{DBLP:conf/naacl/ChenK0CR19} & 20k & EN & Text & Claim Detection \& Evidence Detection & & \ding{51}  \\

UKP ASPECT & \citeyear{DBLP:conf/acl/ReimersSBDSG19} & 3.5K & EN & Text & Claim Detection & & \ding{51} \\

IBM Labeled Emphasized Words in Speech & \citeyear{DBLP:conf/interspeech/MassSMHSLK18} & 2.4k & EN & Multi & Text to Speech & & \ding{51} \\\

AMPERSAND & \citeyear{DBLP:journals/corr/abs-2004-14677} & 2.7k & EN & Text & Argument Relation Identification & & \ding{51} \\

CoPA & \citeyear{DBLP:conf/acl/BiluGHSLMMGS19} & 800 & EN & Text & Claim Detection & & \ding{51} \\

IBM ArgQ-14kArgs & \citeyear{DBLP:conf/emnlp/ToledoGCFVLJAS19} & 14k & EN & Text & Convincingness Checking & & \ding{51} \\

IBM Argumentative Sentences in Recorded Debates & \citeyear{DBLP:conf/emnlp/ShnarchCMAS20} & 700 & EN & Text & Claim Detection & & \ding{51} \\

Review-Rebuttal & \citeyear{DBLP:conf/emnlp/ChengBYLS20} & 4.7K & EN & Text & Claim Detection & & \ding{51} \\

IBM Evidences Sentences & \citeyear{DBLP:conf/aaai/Ein-DorSDHSGAGC20} & 29k & EN & Text & Convincingness Checking & & \ding{51} \\

IBM XArgMining & \citeyear{DBLP:conf/emnlp/Toledo-RonenOBS20} & 35k & DE,NL,E S,FR,IT & Text & Claim Detection \& Evidence Detection \& Stance Detection & & \ding{51} \\

IBM Recorded Debating Dataset & \citeyear{DBLP:conf/acl/OrbachBTLJAS20} & 3.5k & EN & Multi & Speech to Text & & \ding{51} \\

MEDLINE & \citeyear{DBLP:conf/ecai/0002CV20} & 6.7k & EN & Text & Claim Detection \& Argument Relation Identification & & \ding{51} \\

GAQCorpus & \citeyear{DBLP:conf/coling/LauscherNNT20} & 5k & EN & Text & Convincingness Checking \& Sufficiency Checking & & \ding{51} \\

IBM ArgKP & \citeyear{DBLP:conf/argmining/FriedmanDHAKS21} & 27k & EN & Text & Key Point Analysis & & \ding{51} \\

SciARK & \citeyear{DBLP:conf/argmining/FergadisPKP21} & 1k & EN & Text & Claim Detection \& Evidence Detection & & \ding{51} \\

\rowcolor{orange!10}
IAM & \citeyear{DBLP:conf/acl/ChengBHYZS22} & 69k & EN & Text & Claim Detection \& Evidence Detection \& Stance Detection & & \ding{51} \\
\rowcolor{orange!10}
FinArg & \citeyear{alhamzeh2022s} & 136 & EN & Text & Claim Detection \& Stance Detection & &  \\
\rowcolor{orange!10}
Sustainable Diet Arguments on Twitter & \citeyear{hansen2022dataset} & 592 & EN & Text & Claim Detection \& Evidence Detection \& Stance Detection & & \ding{51} \\
    
\bottomrule
\multicolumn{7}{l}{\footnotesize\textit{Modality:} text, audio, video, multi.
‘LLM Origin’ \ding{51} = created \emph{with} or \emph{for} LLM evaluation.}
\end{tabular}

\caption{Recent corpora; shaded rows belong to the “LLM era’’ (2021–present). German (DE), Dutch (NL), Spanish (ES), French (FR), and Italian (IT).}
\label{tab:datasets_statistics_2021}
\end{table*}

%% file: table/dataset_statistics_2025.tex
\begin{table*}[t]
\small\setlength{\tabcolsep}{5pt}

\begin{tabular}{@{}C{2.8cm}C{1cm}C{1cm}C{1cm}C{1cm}C{5cm}C{1.2cm}C{1cm}@{}}
\toprule
\textbf{Dataset} & \textbf{Year} & \textbf{Size} & \textbf{Lang.} &
\textbf{Modality} & \textbf{Main Tasks} & \textbf{LLM Origin\footnotemark} & \textbf{Open Source}\\
\midrule

Kialo & \citeyear{agarwal2022graphnli} & 310k & EN & Text & Argument Relation Identification & & \ding{51} \\

IBM SurveyKP & \citeyear{DBLP:conf/emnlp/EdenKOKSB23} & 15k & EN & Text & Key Point Analysis & & \ding{51} \\

ArguGPT & \citeyear{DBLP:journals/corr/abs-2304-07666} & 8k & EN & Text & Claim Detection & \ding{51} & \ding{51} \\

StoryARG & \citeyear{DBLP:conf/acl/FalkL23} & 2.4k & EN & Text & Claim Detection \& Argument Relation Identification &  & \ding{51} \\

GENEVA & \citeyear{DBLP:conf/acl/ParekhHHCP23} & 3k & EN & Text &  Argument Relation Identification &  & \ding{51} \\

VivesDebate-Speech & \citeyear{DBLP:conf/emnlp/Ruiz-DolzS23} & 29 & EN & Multi &
Speech to Text &  & \ding{51} \\  

ArgSciChat & \citeyear{DBLP:conf/acl/RuggeriMG23} & 498 & EN & Text &   Argument Relation Identification &  & \ding{51} \\

MuLMS-AZ & \citeyear{DBLP:journals/corr/abs-2307-02340} & 10k & EN & Text &   Argument Relation Identification &  & \ding{51} \\

ArgAnalysis35K & \citeyear{DBLP:conf/acl/JoshiPH23} & 35k & EN & Text &   Argument Relation Identification &  &  \\

ORCHID & \citeyear{DBLP:conf/emnlp/ZhaoWP23} & 14k & EN & Text & Argument Summarization \& Stance Detection  &  & \ding{51} \\

ARIES & \citeyear{gemechu2024aries} & 81k & EN & Text & Argument Relation Identification & & \ding{51} \\

FS150T Corpus & \citeyear{DBLP:conf/emnlp/SchillerDWG24} & 21k & EN & Text & Claim Detection \& Stance Detection & & \ding{51} \\

CasiMedicos-Arg & \citeyear{DBLP:conf/emnlp/SviridovaYECVA24} & 7k & EN & Text & Claim Detection \& Stance Detection & & \ding{51} \\

CEAMC & \citeyear{DBLP:conf/emnlp/RenWLZZYZBL24} & 226 & EN,ZH & Text & Claim Detection \& Evidence Detection \& Stance Detection & & \ding{51} \\

ASE &
\citeyear{DBLP:conf/acl/0074WSBMZZWHLN24} & 2.3 k & EN & Text &
Evidence Detection \& Convincingness Checking \& Argument Summarization \& Sufficiency Checking &
\ding{51} & \ding{51} \\  

ConQRet & \citeyear{DBLP:conf/naacl/DholeSA25} & 6.5k & EN & Text &
Evidence Detection \& Argument Quality Assessment & \ding{51} & \ding{51} \\

\bottomrule
\multicolumn{7}{l}{\footnotesize\textit{Modality:} text, audio, video, multi.
‘LLM Origin’ \ding{51} = created \emph{with} or \emph{for} LLM evaluation.}
\end{tabular}

\caption{Continue from previous table. Recent corpora belong to the “LLM era’’ (2021–present). German (DE), Dutch (NL), Spanish (ES), French (FR), and Italian (IT). Chinese (ZH)}
\label{tab:datasets_statistics_2025}
\end{table*}

%% file: section/3_task.tex
\section{Task Taxonomy}
\label{sec:taxonomy}

\begin{figure*}
    \centering
    \includegraphics[width=\linewidth]{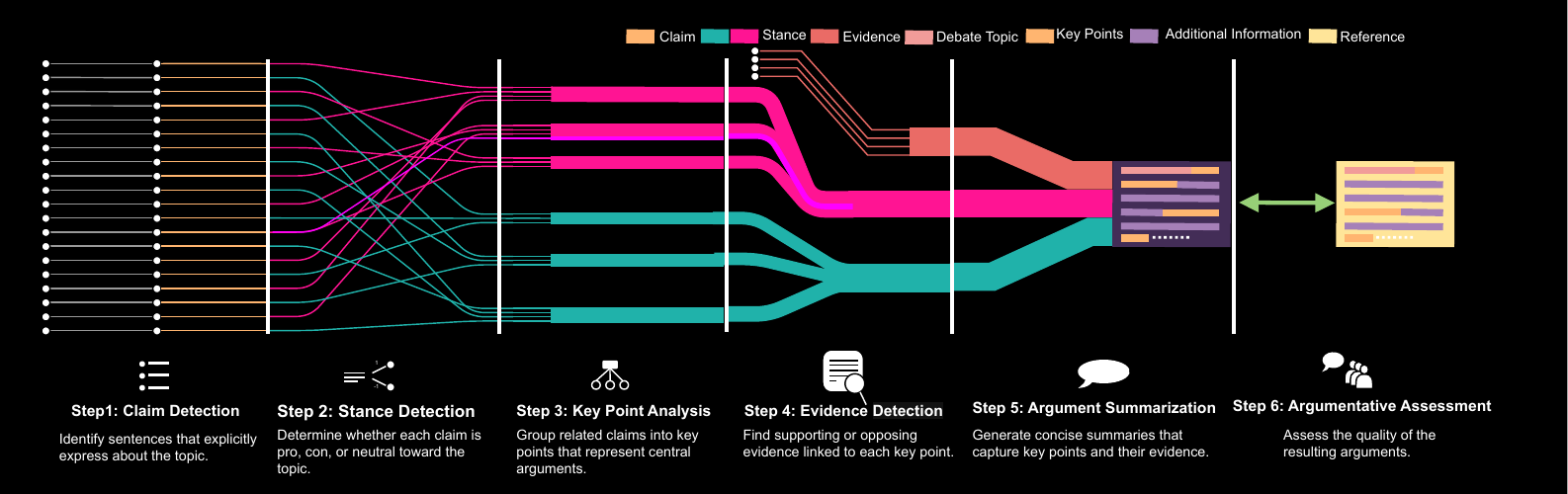}
    \caption{End-to-end workflow for constructing debate-ready argument summaries. This pipeline illustrates the hierarchical progression from Step 1: Claim Detection, where sentence-level argumentative units are identified, through Step 2: Stance Detection and Step 3: Key Point Analysis, which organize claims by polarity and cluster them into coherent key points. Step 4: Evidence Detection retrieves supporting or opposing evidence for each key point, enabling Step 5: Argument Summarization to generate concise, structured argumentative summaries. Finally, Step 6: Argumentative Assessment evaluates the completeness, coherence, and quality of the produced arguments. The Sankey-style flow visualizes how claims, stances, evidence, and key points propagate through the pipeline to form coherent debate scripts.}
    \label{fig:task_pipeline}
\end{figure*}

\subsection{Claim Detection}

Claim detection constitutes the foundational step of argument mining by identifying propositions that express a stance, assert factuality, or require further justification. Prior work spans a wide range of methodological paradigms and domain settings. To provide conceptual clarity, we organise this literature along three axes: (1) methodological developments, (2) application-driven domains, and (3) multilingual and multi-modal extensions. 

\subsubsection{Methodological Dimensions}

Early claim detection methods relied on rule-based extraction and heuristic templates, which proved limited in generalisation and cross-domain transferability. Contemporary approaches increasingly adopt neural architectures, distant supervision, and multimodal reasoning.

\paragraph{Sequence Labelling and Transformer-Based Classification.}
A line of work models claim detection as a span classification or token-labelling task. Transformer-based classifiers, particularly those fine-tuned on domain-specific datasets, have significantly improved robustness across noisy or informal environments. Systems such as ClaimHunter \citep{DBLP:conf/www/BeltranML21} and FactRank \citep{DBLP:journals/osnm/BerendtBHJPA21} exemplify this trend by deploying contextual encoders for real-time claim identification on social platforms. More recent frameworks such as CLAIMSCAN \citep{DBLP:conf/fire/SundriyalA023} leverage contextual signals to detect claim spans and misinformation indicators.

\paragraph{Weak Supervision and PU-Learning.}
Weakly supervised and positive–unlabeled learning techniques address annotation scarcity in dynamic contexts such as political discourse. \citet{DBLP:conf/emnlp/WrightA20} demonstrate that PU-learning improves claim-level classification in high-noise environments. Fairness-aware detection approaches, such as actor masking \citep{DBLP:conf/acl/DayanikP20}, further mitigate model bias toward politically salient entities.

\paragraph{LLM-Based and Instruction-Following Approaches.}
Instruction-tuned LLMs have recently been explored for claim detection across languages and modalities. Studies such as \citet{DBLP:journals/corr/abs-2503-02737} and \citet{DBLP:journals/corr/abs-2503-15220} show that large generative models can transfer claim detection capabilities with minimal supervision, although cross-lingual semantic drift remains a key challenge. These works highlight the emerging shift toward zero-shot and few-shot claim extraction grounded in broad world knowledge.

\subsubsection{Application Domains}

The linguistic and argumentative structure of claims varies substantially across domains. Existing research thus develops specialised datasets and modeling strategies to address domain-specific complexity.

\paragraph{Social Media and Web Platforms.}
Social media environments present substantial challenges due to short, informal, and fast-evolving discourse. Beyond ClaimHunter and FactRank, multimodal integration has proven beneficial: MM-Claims \citep{DBLP:conf/naacl/CheemaHSMOE22} introduces a dataset integrating textual and visual evidence to improve claim understanding. Additional frameworks such as CLAIMSCAN \citep{DBLP:conf/fire/SundriyalA023} focus on span-level truth discovery in noisy digital environments.

\paragraph{Healthcare and Biomedical Claims.}
Biomedical claim detection examines health-related assertions, contradictory scientific reporting, and public health misinformation. Prior work includes fraud identification \cite{DBLP:journals/ijcse/GaoGG19}, blockchain-secured verification systems \citep{DBLP:conf/icacds/MohanP19}, and optimisation-based frameworks \citep{DBLP:journals/kbs/TubishatTAHH25}. Studies such as \citet{DBLP:conf/bionlp/WuhrlK21} focus on health claims within Twitter discourse, while \citet{DBLP:conf/infrkm/YaziVRTL21} examine contradictory biomedical findings. A recent systematic review \citep{DBLP:journals/artmed/PreezBBB25} synthesises machine learning approaches to healthcare fraud and behavioural anomalies.

\paragraph{Environmental and Sustainability Claims.}
Greenwashing detection and environmental fact-checking require specialised treatment due to scientific terminology and organisation-specific reporting structures. Prior work includes domain-specific sustainability claim detectors \citep{DBLP:conf/fire/WoloszynKS21} and broader frameworks addressing climate misinformation \citep{DBLP:conf/acl/StammbachWBKL23}.

\paragraph{Financial and Numerical Claims.}
Financial communication contains quantitative assertions, forecasts, and performance claims. NumClaim \citep{DBLP:conf/cikm/ChenHC20} detects numerical inconsistencies in investor-oriented texts, while new datasets \citep{DBLP:journals/corr/abs-2402-11728} support weakly supervised extraction of claims from earnings reports and financial analyses.

\paragraph{Educational and Scientific Discourse.}
Claim detection has also been applied to writing assessment and scientific discourse interpretation. \citet{DBLP:conf/edm/WanCAM20} show that claim frequency correlates with academic writing quality, whereas \citet{DBLP:phd/hal/Hafid24} analyse scientific claims within online discussions and scholarly communities.

\paragraph{Political Fact-Checking and Rumour Detection.}
High-stakes political contexts motivate robust and fairness-aware claim detection. Actor-masked training \citep{DBLP:conf/acl/DayanikP20}, PU-learning \citep{DBLP:conf/emnlp/WrightA20}, and adversarial robustness studies \citep{DBLP:journals/corr/abs-2407-18367} address bias and manipulation. Rumour detection architectures such as claim-guided hierarchical attention networks \citep{DBLP:conf/emnlp/LinMCYCC21} further highlight the interaction between claim signals and misinformation propagation.

\subsubsection{Multilingual and Cross-Lingual Claim Detection}

Multilingual misinformation ecosystems have motivated research on cross-lingual transfer and low-resource claim detection.

\paragraph{Adapter Fusion and Cross-Lingual Transfer.}
\citet{DBLP:conf/ecir/SchlichtFR23} demonstrate that adapter fusion enables efficient multilingual transfer for claim detection in non-English settings. Complementary work \citep{DBLP:journals/corr/abs-2503-15220} investigates knowledge transfer between linguistically distant languages, illustrating the difficulty of aligning semantic structures across global misinformation narratives.

\paragraph{LLM-Based Multilingual Detection.}
Multilingual LLMs \citep{DBLP:journals/corr/abs-2503-02737} provide strong zero-shot capabilities but require careful handling of cultural references and latent biases. These works collectively reveal that cross-lingual generalisation remains limited without domain-aligned adaptation.

\subsubsection{Multimodal Claim Detection}

Claims increasingly appear in contexts combining text, images, or video. This has led to the development of multimodal datasets and architectures.

\paragraph{Image to Text Claims.}
MM-Claims \citep{DBLP:conf/naacl/CheemaHSMOE22} demonstrates that incorporating visual features improves robustness to misleading or manipulated imagery that alters claim semantics.

\paragraph{Video and Audiovisual Content.}
Recent efforts extend claim detection to video-based misinformation. \citet{DBLP:conf/visigrapp/Rayar24a} explore multimodal video analysis, while ViClaim \citep{DBLP:journals/corr/abs-2504-12882} provides a multilingual benchmark integrating audiovisual and textual evidence. These settings present additional challenges including speech-to-text noise, temporal alignment, and cross-modal grounding.

\subsubsection{Review and Challenges}

Although claim detection has matured across several methodological and application-driven dimensions, substantial gaps remain that limit the reliability and generalisability of current systems. A persistent challenge concerns the detection of implicit and context-dependent claims, particularly in domains where argumentative intent is subtle, distributed across sentences, or shaped by multimodal cues. Existing systems are effective for explicit, clause-level claims, yet often fail when claims are implied through presupposition, metaphor, or narrative framing. This limitation reflects a deeper issue: current benchmarks overwhelmingly favour well-formed, explicit annotations, leaving implicit claim phenomena systematically underrepresented in evaluation.

Domain generalisation represents another central gap. Many approaches perform well on static, topic-specific datasets but degrade significantly in dynamic discourse such as political communication or public health misinformation. These contexts evolve rapidly, require temporal awareness, and incorporate strategic rhetorical shifts intended to evade detection. Current transformer-based models excel at local semantic classification but struggle to maintain robustness under topic drift, adversarial reframing, or hybrid misinformation strategies. Achieving robustness across time, domains, and discourse communities remains a critical but unresolved objective.

Cross-lingual claim detection further illustrates the tension between model capacity and linguistic diversity. While multilingual transformers and LLM-based approaches demonstrate promising transfer capabilities, they remain sensitive to cultural references, idiomatic expression, and region-specific political or scientific knowledge. These shortcomings are amplified in low-resource languages lacking annotated corpora. The field currently lacks principled evaluation protocols for assessing semantic fidelity across languages, hindering our ability to quantify transfer degradation and identify where cross-lingual misalignment emerges.

Finally, multimodal claim detection, despite rapid progress, faces substantial grounding and alignment challenges. Claims expressed in image–text or video–text formats often derive meaning from the interaction between modalities, requiring models to integrate visual context, temporal structure, and discourse-level information. Existing approaches capture correlations but rarely achieve genuine multimodal grounding, making them susceptible to visually incongruent or deliberately manipulated content. As multimodal misinformation becomes more sophisticated, the need for models capable of reasoning jointly over modality-specific evidence and argumentative structure becomes increasingly urgent.

\subsection{Stance Detection}

Stance detection aims to identify the position expressed by an author toward a given target, claim, or topic, most commonly categorised as \textit{pro}, \textit{con}, or \textit{neutral}. Unlike sentiment analysis, stance detection is inherently target-dependent and frequently requires implicit reasoning, contextual interpretation, and understanding of discourse relations. This task plays a central role in argument mining, political analysis, misinformation studies, and social media research. In this section, we organise stance detection research into its problem formulations, datasets, methodological developments, generalisation challenges, bias mitigation strategies, applications, and data augmentation techniques.

\subsubsection{Neural and Classic Approaches}

Early neural stance detection models relied on BiLSTMs or attention-based architectures.  
\citet{DBLP:conf/sbp-brims/HosseiniaDM19} employ BiLSTM models with topic-aware representations to capture argumentative stance.  
More recent work incorporates probabilistic reasoning and topic modelling: \citet{DBLP:journals/mta/WangPW24} introduce a recurrent stick-breaking topic model that jointly models stance and latent topical structure.  
Social influence and propagation cues have also been explored: \citet{DBLP:journals/concurrency/LiYWW24} simulate rumour diffusion processes to improve stance prediction in socially networked environments.  
These methods demonstrate that traditional stance detection benefits from both linguistic and contextual signals.

\subsubsection{LLM-Based and Reasoning-Enhanced Approaches}

With the emergence of large language models, stance detection has increasingly incorporated zero-shot prompting, rationale-guided modeling, and instruction tuning.  
\citet{DBLP:conf/acl/ZhaoLCZ24} propose \textbf{ZeroStance}, using ChatGPT to generate stance-annotated datasets and perform zero-shot stance detection.  
\citet{DBLP:journals/ml/GambiniSFT24} systematically evaluate LLMs such as GPT-3 and BLOOM on stance benchmarks, showing that LLMs capture implicit stance surprisingly well but exhibit instability under target shifts.  
Rationale-based modeling has also emerged: LOGIC \citep{DBLP:journals/peerj-cs/LeeLK24} uses LLM-generated rationales to guide smaller models, improving stance reasoning transparency.  
These approaches highlight LLMs' strong reasoning capacity while exposing challenges in calibration, bias, and target dependency.

\subsubsection{Generalisation: Cross-Lingual, Cross-Domain, Zero-Shot, and Few-Shot}

Stance detection must generalise across topics, cultures, and linguistic contexts.  
Multilingual efforts such as those by \citet{DBLP:journals/snam/CharfiBAAAZ24} and \citet{DBLP:conf/wanlp/AlghaslanA24, DBLP:conf/wanlp/ShuklaVK24} build Arabic corpora and leverage LLM fine-tuning or ensembling to improve multilingual stance robustness.  
Zero-shot and few-shot generalisation has received increasing attention.
Chain-of-Stance \citep{DBLP:conf/nlpcc/MaWXZZ24, DBLP:journals/corr/abs-2408-04649} uses chain-of-thought reasoning to infer stance toward unseen targets, while \citet{DBLP:conf/icic/GuoJL24} incorporate structured external knowledge to improve generalisation under target shifts.  
Adversarial learning techniques such as those by \citet{DBLP:journals/ijon/ZhangLZL24} integrate commonsense knowledge to stabilise stance predictions in challenging contexts.  
These works collectively suggest that stance detection generalisation remains an open challenge requiring reasoning, abstraction, and cultural grounding.

\subsubsection{Bias Mitigation and Calibration}

LLMs exhibit ideological, demographic, and cultural biases that directly influence stance predictions.  
\citet{DBLP:conf/naacl/LiZLGWZLWX25} introduce counterfactual augmented calibration to reduce stance-related biases by perturbing target-specific lexical cues.  
\citet{DBLP:journals/corr/abs-2402-14296} further examine bias calibration strategies to stabilise stance outputs across demographic or topic variations.  
These studies underscore the risk that stance detectors—especially LLM-based ones—can inherit or amplify societal biases, making bias mitigation an essential direction for future research.

\subsubsection{Applications Across Domains}

Stance detection is widely applied in political analysis, online debates, and health misinformation.  
In public health, \citet{DBLP:journals/jbi/DavydovaYT24} analyse stance toward COVID-19 on social media, while \citet{DBLP:journals/corr/abs-2411-14720} study HPV vaccine stance using fine-tuned and in-context LLMs.  
Open-target stance detection for health communication is explored in \citet{DBLP:journals/corr/abs-2409-00222}, demonstrating that stance identification can support misinformation monitoring and epidemiological analysis.  
These applications demonstrate the breadth of stance detection and its importance in high-stakes social and public health domains.





\subsubsection{Review and Challenges}

Stance detection has evolved into a rich research area spanning neural models, LLM-based reasoning, multilingual adaptation, and synthetic data generation. Yet significant challenges remain. A central unresolved issue is the detection of \emph{implicit} stance, where the author's position is encoded through narrative framing, presupposition, or domain knowledge rather than explicit markers. Models excel at explicit stance classification but remain brittle under subtle or culturally dependent cues. Generalisation across topics, targets, and linguistic communities is another persistent difficulty; while LLMs improve zero-shot performance, they also introduce instability and ideological bias, particularly when targets shift or new sociopolitical contexts emerge.

Moreover, multi-turn and conversational stance detection remains underexplored. Real-world stance often develops over time, requiring models to track discourse dynamics, pragmatic shifts, and historical context. Existing datasets cover some aspects of stance evolution, yet current models typically ignore temporal structure. Multi-modal stance detection also requires further exploration, especially as political and health communication increasingly leverage images and videos. Finally, efforts in bias mitigation and calibration highlight the need for principled frameworks ensuring fairness and transparency. Together, these challenges indicate that future advances will depend on deeper contextual reasoning, robust cross-domain representations, culturally grounded evaluation, and models capable of capturing the complexity of human argumentative stance.

\subsection{Key Point Analysis}
Key Point Analysis (KPA) is a structured framework for identifying concise, recurring, and representative ideas—referred to as key points—from large collections of arguments. Originally introduced by \citet{DBLP:conf/acl/Bar-HaimEFKLS20}, KPA addresses limitations of free-form argument summarization by enabling interpretable, reusable, and scalable representations of public discourse. Over time, KPA has evolved into a multi-faceted research area encompassing supervised extraction, semantic matching, contrastive learning, and more recently, LLM-driven generation and abstraction.

\subsubsection{Key Point Matching}

Early work on KPA established the core paradigm of extracting key points and matching arguments to them. \citet{DBLP:conf/acl/Bar-HaimEFKLS20, DBLP:conf/acl/Bar-HaimEKFS20} introduced the first supervised and unsupervised methods, framing key point matching as a semantic similarity task. These contributions motivated a series of shared tasks and benchmark datasets that standardized evaluation protocols for KPA. Subsequent research explored lightweight yet effective models. \citet{DBLP:conf/argmining/PhanNND21} proposed a bi-encoder architecture that achieved strong performance with minimal computational overhead. \citet{DBLP:conf/argmining/ReimerLHA21} further examined how pretrained language models could enhance both key point extraction and alignment, highlighting the importance of representation quality.

To improve the robustness and generalization of key point matching, researchers explored richer representational frameworks. Contrastive learning emerged as a powerful strategy. \citet{DBLP:conf/argmining/AlshomaryGSHSCP21} demonstrated that contrastive objectives can better separate fine-grained argumentative distinctions, especially in noisy or overlapping opinion spaces. Graph-based techniques also played a substantial role. \citet{DBLP:conf/naacl/LiJHXCH24} introduced graph representations to capture relational structures between arguments and candidate key points, enabling more context-aware and structurally coherent matching. These advances collectively improved scalability to real-world argumentative corpora. Real-world deployment needs spurred incremental and dynamic systems. \citet{DBLP:conf/emnlp/EdenKOKSB23} developed an online KPA framework capable of updating key points as new arguments emerge, addressing the evolving nature of public opinions.

\subsubsection{Key Point Generation}

The rise of large language models has reshaped KPA, expanding the task beyond rigid matching toward more generative and adaptive paradigms. Several works integrated LLM-based prompting, abstraction, and iterative refinement. LLMs have been used to generate aspect-specific or domain-adapted key points. \citet{DBLP:conf/acl/Tang0DC24, DBLP:conf/eacl/TangZD24} explored prompted KPA in review summarization, showing that LLMs can learn fine-grained aspects with limited supervision. \citet{DBLP:conf/acl/LiSBN23} proposed an iterative clustering–summarization pipeline that combines LLM abstraction with clustering to automatically induce high-quality key points. More recently, \citet{DBLP:journals/corr/abs-2404-18371} demonstrated that zero-shot KPA is feasible through question generation and network analysis, suggesting that LLMs possess latent capabilities for structuring opinion spaces without explicit training.

\subsubsection{Review and Challenges}

Key Point Analysis has developed into a versatile and increasingly mature subfield of argument summarization, integrating insights from clustering, semantic similarity modeling, contrastive representation learning, and generative LLM reasoning. Yet, several research gaps remain. Current systems still struggle with identifying key points that are both broadly representative and sufficiently fine-grained to capture minority perspectives. The balance between abstraction and fidelity remains an open question—especially in LLM-generated key points, which can introduce hallucinated or overly generic abstractions. Moreover, domain adaptation continues to be a bottleneck: KPA models trained on public argument datasets can degrade significantly when applied to specialized domains such as healthcare, law, or scientific discourse. Finally, scalability challenges—particularly in continuously evolving online environments—demand more dynamic and self-updating KPA frameworks that maintain stability while adapting to shifting argument distributions. These challenges present opportunities for integrating retrieval-augmented generation, structured LLM reasoning, and hybrid symbolic–neural architectures to push the boundaries of key point quality, reliability, and real-world applicability.

\subsection{Evidence Detection}

Evidence detection is a core component of argument mining. It involves identifying textual, cross-lingual, or multimodal elements that support or refute a claim. Unlike claim detection, which focuses on recognising the proposition itself, evidence detection aims to locate the information necessary for validation, verification, or contradiction resolution. This task has expanded significantly across domains such as fake news detection, political fact-checking, scientific discourse analysis, and multimodal misinformation detection. To provide conceptual clarity, this section organises prior work into methodological formulations, neural and graph-based approaches, iterative retrieval techniques, multilingual and cross-domain settings, multimodal evidence grounding, and contradiction-focused evidence analysis.



\subsubsection{Pretrained Language Model Approaches}

Pretrained language models (PLMs) and LLM-based frameworks have become central to contemporary evidence detection. Dementieva and Panchenko \citep{DBLP:conf/acl/DementievaP21, DBLP:conf/dsaa/DementievaP20} demonstrate the utility of cross-lingual textual evidence for strengthening monolingual misinformation detection, showing that multilingual contextual signals significantly enhance evidence extraction quality. Self-trained PLMs fine-tuned specifically for evidence detection have also shown notable improvements: for instance, \citet{DBLP:conf/argmining/ElarabyL21} employ domain-adaptive PLMs to boost evidence selection accuracy in argumentative contexts.

\subsubsection{Graph-Based and Structured Evidence Modeling}

Beyond PLM-centric models, graph-based structures have played an increasingly important role in modeling multi-source or cross-document evidence. Evidence graphs capture relational dependencies between claims, sources, and supporting statements. The work of \citet{DBLP:conf/www/XuWLWW22, DBLP:journals/corr/abs-2201-06885} exemplifies fine-grained semantic graph mining, while interpretable structures such as those proposed by Guo et al. \citet{DBLP:conf/cikm/GuoZT023} demonstrate the benefit of structured reasoning pipelines. These systems move beyond simple retrieval toward evidence aggregation and coherence modeling.

\subsubsection{Iterative and Multi-Step Evidence Retrieval}

Evidence detection often requires reasoning through multiple steps. Iterative retrieval frameworks explicitly model the sequential nature of evidence acquisition. Liao et al. introduce MUSER \citep{DBLP:conf/kdd/LiaoPHZLSX23, DBLP:journals/corr/abs-2306-13450}, a multi-step retrieval enhancement system that iteratively refines retrieved evidence, significantly improving performance in complex misinformation settings. \citet{DBLP:conf/ijcai/WuWZ24} propose a unified evidence enhancement inference architecture capable of dynamically incorporating new information during prediction. These approaches highlight the importance of multi-hop, feedback-driven retrieval for complex or incomplete evidence environments.

\subsubsection{Multilingual Evidence Detection}

Multilingual settings introduce additional complexity due to linguistic divergence and culturally specific information.  \citet{DBLP:journals/jimaging/DementievaKP23} propose Multiverse, a multilingual evidence detection framework that integrates semantically aligned evidence across diverse linguistic contexts. Similarly, \citet{DBLP:journals/access/HammouchiG22} design multilingual misinformation detection systems that explicitly incorporate evidence-aware representations, demonstrating the importance of cross-lingual knowledge transfer. These approaches confirm that multilingual evidence is crucial for robust global misinformation detection.

\subsubsection{Cross-Domain and Cross-Platform Evidence Detection}

Evidence detection systems often generalise poorly when encountering distributional shifts or unseen misinformation types.  \citet{DBLP:journals/tkde/LiuWWW24} address this challenge through dual adversarial debiasing, mitigating out-of-distribution degradation in evidence-aware models.  \citet{DBLP:journals/fi/FerdushKKGD25} highlight the importance of cross-platform evidence integration, showing that evidence derived from multiple social media platforms improves robustness and transferability. These works emphasise the need for domain-stable evidence representations.

\subsubsection{Multimodal Evidence Detection}

Evidence frequently spans modalities, especially in digital misinformation contexts. \citet{DBLP:journals/eaai/HuangMRJS25} introduce dual evidence enhancement mechanisms integrating text–image similarity signals, demonstrating that multimodal consistency plays a key role in misinformation verification. \citet{DBLP:conf/icassp/WuC24} further develop an evidence-aware multimodal model tailored for Chinese social media, effectively capturing culturally specific multimodal cues. These studies collectively illustrate that multimodal alignment and cross-modal corroboration are central challenges in modern evidence detection.

\subsubsection{Counter-Evidence and Contradiction Detection}

Beyond supportive evidence, systems must also identify conflicting, contradictory, or refuting evidence. \citet{DBLP:conf/emnlp/JiayangCZQZLS0L24} propose the ECON framework, which focuses on detecting and resolving evidence conflicts—an essential capability for factual integrity and claim verification.  \citet{DBLP:conf/iccci/KharratHR22} further examine contradiction detection by integrating semantic relational cues with uncertainty estimation. Explainability also remains crucial: \citet{DBLP:journals/corr/abs-2407-01213} introduce EMIF, a multi-source evidence fusion network that emphasises interpretability and transparent evidence grounding. Together, these studies highlight the central role of conflict-aware reasoning in robust evidence detection.

\subsubsection{Review and Challenges}

Evidence detection has advanced substantially through pretrained language models, graph-based reasoning, iterative retrieval frameworks, and multilingual or multimodal extensions. Yet important gaps remain. One persistent challenge concerns the inherently distributed nature of evidence: supporting or refuting information often spans multiple documents, modalities, or discourse contexts, making retrieval a multi-hop and multi-source reasoning problem. Although iterative retrieval systems have improved evidence accumulation, they still struggle with implicit, incomplete, or weakly aligned evidence.

Generalisation also remains a major barrier. Many evidence detection models perform well within curated benchmarks but degrade under distribution shifts such as emerging misinformation events, evolving rhetorical strategies, or new platforms. Cross-domain robustness requires systems to disentangle domain-specific surface features from deeper evidential structures—a capability that current PLM-based models only partially achieve.

Multilingual and cross-cultural evidence introduces another layer of difficulty. Evidence expressed in different languages may encode culturally specific knowledge, implicit assumptions, or contextual signals that multilingual models fail to capture. Existing multilingual systems rely heavily on machine translation or alignment-based transfer, both of which risk semantic drift. More principled frameworks for cross-lingual evidential reasoning remain scarce.

Lastly, multimodal and contradictory evidence pose increasing complexities. Modern misinformation campaigns frequently rely on visuals, modified images, or videos that subtly shift meaning. While recent multimodal approaches improve text–image consistency modeling, fully grounded multimodal reasoning remains an open challenge. Similarly, identifying contradictory or conflicting evidence requires models to integrate uncertainty estimation, logical reasoning, and semantic nuance—capabilities not yet fully supported by current architectures.

Addressing these challenges will require richer benchmarks, deeper integration of retrieval and reasoning, and principled models capable of understanding how evidence is constructed, distributed, and contested across languages, modalities, platforms, and domains.

\subsection{Argument Summarization}

Argument summarization aims to generate concise, faithful, and discourse-aware summaries of argumentative content, including debates, deliberative dialogues, legal opinions, and online discussions. Unlike generic text summarization, argumentative summarization must preserve key claims, supporting or opposing evidence, stance polarity, and the interaction between argumentative units. This makes the task structurally complex and methodologically heterogeneous, requiring coordinated treatment of content selection, discourse modeling, and abstraction.

\subsubsection{Modeling Paradigms and Learning Frameworks}

A range of modeling approaches has been explored to handle argumentative structure. Early work established the need to integrate argument mining signals into the summarization pipeline. ConvoSumm \citep{DBLP:conf/acl/FabbriRRWLMR20} demonstrated that identifying argumentative components (claims, premises, and rebuttals) substantially improves the informativeness and structure of dialogue summaries. This line of research clarified that argument summarization is not merely a compression task but an interpretive process that must map a set of dialogic or adversarial exchanges into a coherent argumentative representation. In legal summarization, models such as ArgLegalSumm \citep{DBLP:conf/coling/ElarabyL22} incorporate argument mining predictions into abstraction pipelines, allowing systems to retain central legal claims and supporting facts. Reranking extensions proposed by \citet{DBLP:conf/acl/ElarabyZL23} further improve coherence and informativeness by selecting summaries aligned with argumentative salience.

Beyond domain-specific architectures, several approaches explicitly model argumentative relations. Shan and Lu \citep{DBLP:journals/ijon/ShanL25} introduce a multi-task framework combining summarization with argument-relation–aware contrastive learning, enabling models to better capture divergence and opposition in debates. Syed et al.\ \citep{DBLP:conf/sigdial/SyedZHWP23} shift the focus to frame-oriented summarization, capturing pragmatic and rhetorical dimensions often overlooked by content-based approaches. Complementary evaluations by Meer et al.\ \citep{DBLP:conf/eacl/MeerVJM24} examine the diversity of generated summaries and highlight persistent trade-offs between informativeness and representation of multiple perspectives.

Recently, iterative generative paradigms—especially diffusion-based models—have gained traction. Their ability to revise intermediate states makes them particularly suitable for argument summarization, where preserving logical consistency and multi-claim structure is essential. \citet{li2025arg} demonstrate that diffusion models can refine argumentative content through multi-step denoising, leading to more coherent and balanced summaries.

\subsubsection{Domain-Specific Adaptations}
Argument summarization has expanded into specialized domains where argumentative structure is long-form, technical, or high-stakes. Legal summarization constitutes a prominent branch, with multiple works addressing the extraction of legal reasoning \citep{DBLP:conf/acl/ElarabyZL23, DBLP:conf/coling/ElarabyL22, DBLP:conf/icail/XuA23, DBLP:conf/jurix/SteffesR22}. These systems must manage complex rhetorical moves such as statutory interpretation, case comparisons, and multi-level evidence chains, pushing summarization models toward richer structural encoding.


\subsubsection{LLMs Paradigms}
Large language models (LLMs) have redefined argument summarization by enabling reasoning-driven abstraction, multi-step content reorganization, and implicit discourse modeling. Their strong performance on argumentative datasets suggests that LLMs implicitly encode argumentation patterns and pragmatic cues, though their behavior remains sensitive to prompting and domain shifts. Recent research also couples LLMs with retrieval, planning, or iterative refinement modules, suggesting a trend toward hybrid symbolic–neural summarization workflows \citep{DBLP:conf/emnlp/ZhaoWP23, DBLP:conf/nips/RoushSBZMZBVS24}.

\subsubsection{Review and Challenges}
Argument summarization has evolved from early discourse-aware extractive models into a sophisticated generative task requiring multi-layered reasoning. However, significant challenges persist. Current systems still struggle with representational fidelity, especially in adversarial or multi-stance settings where preserving argumentative balance is critical. While LLMs introduce powerful generalization capabilities, their summarization often reflects stylistic homogenization, insufficient evidence grounding, or implicit biases inherited from pretraining data. Domain-specific applications such as legal or civic discourse expose limitations in handling long-context dependencies and fine-grained argument structure. Future research must therefore pursue deeper modeling of argument structure, improved interpretability, and robust cross-domain generalization, while integrating emerging paradigms such as diffusion-based iterative refinement and retrieval-assisted summarization.

%% file: section/5_evaluation.tex
\section{Argument Quality Assessment}

\label{sec:evaluation}

\begin{figure*}[t]
    \centering
    \includegraphics[width=\linewidth]{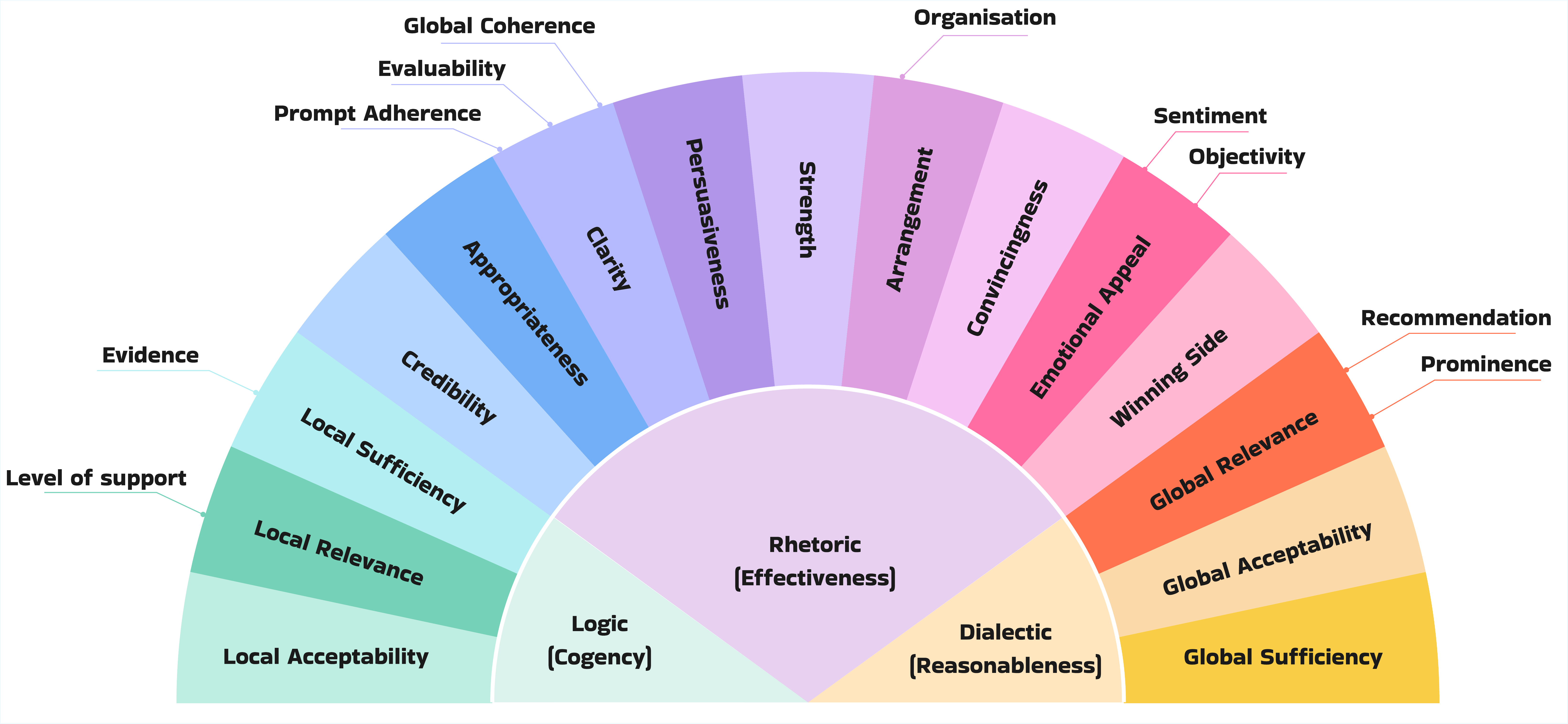}
    \caption{Overview of quality dimensions for argument mining quality assessment discovered in the surveyed literature. Note: The figure is an extension of a taxonomy proposed in \citet{DBLP:conf/emnlp/IvanovaHN24}}
    \label{fig:quality_evaluation_dimension}
\end{figure*}

\emph{Argument Quality Assessment} (AQA)—also termed \emph{Argument Evaluation} or \emph{Argumentation Quality} (AQ), is the stage of the argument-mining pipeline that quantifies \textit{how well} an already-identified argument fulfils its inferential and communicative objectives.  Whereas argument  component identification and relation detection answer the structural questions \textit{“what is an argument and how is it built?”}, AQA addresses the normative question \textit{“how good is this argument?”}.  Early philosophical \citep{wenzel1990three} syntheses distinguish three overarching dimensions, \textbf{Logic/Cogency}, \textbf{Dialectic/Reasonableness}, and \textbf{Rhetoric/Effectiveness}, each further decomposed into sub-dimensions such as acceptability, sufficiency, relevance, audience adaptation, or stylistic appropriateness (shown in Figure \ref{fig:quality_evaluation_dimension}) . 

AQA interacts tightly with but is distinct from argument component detection, relation classification, and generation.  In retrieval settings, quality scores serve as features to re-rank argumentative passages.  In writing-support and educational tools, they trigger targeted feedback on clarity or persuasiveness.  Within persuasive dialogue systems, real-time quality estimates guide strategy selection and counter-argument generation.  Finally, when large language models (LLMs) act both as \emph{generators} and \emph{judges}, AQA becomes a meta-evaluation task, raising new challenges of bias amplification and self-consistency.

\paragraph{Task Definition.}
Formally, given a textual unit $a$, ranging from a single premise to a full argumentative thread, and, optionally, contextual information $c$ (topic, dialogue history, background knowledge), AQA seeks to output either
\emph{(i)} a \emph{scalar} or \emph{categorical} quality score $q(a,c)\in\mathbb{R}$,
\emph{(ii)}  a \emph{vector} of dimension-specific scores $\mathbf{q}(a,c)$, or
\emph{(iii)} a \emph{preference relation} $a \succ b$ between pairs $(a,b)$.
Approaches thus fall along two orthogonal axes:
\emph{absolute vs.\ relative} scoring—Likert-style ratings versus pairwise preferences, and
\emph{intrinsic vs.\ extrinsic} evaluation—judging the argument in isolation versus with external facts or opponent turns .  Each design choice trades annotation complexity for construct validity: relative comparisons often raise inter-annotator agreement, yet absolute scales enable downstream aggregation and benchmarking.

The remainder of this section dissects AQA along five research threads: theoretical quality dimensions (Section \ref{quality_dimensions}), modelling advances from feature engineering to LLM-based judges (Section \ref{llm_evaluation}), evaluation metrics and meta-evaluation (Section \ref{evaluation_metrics}), and applications (Section \ref{evaluation_application}). 

\input{table/quality_dimensions}

\subsection{Quality Dimensions and Theoretical Frameworks}
\label{quality_dimensions}

Argument quality is intrinsically multidimensional.  During the last decade, two complementary lines of work have shaped the field: (i) \emph{theoretical systematisation}, rooted in philosophy and argumentation studies \citep{DBLP:conf/coling/WachsmuthW20}; and (ii) \emph{empirical operationalisation}, driven by the need to annotate large corpora and to train machine-learning models. We synthesise both strands in three steps.

\subsubsection{Classical Tripartite Perspectives: Logic, Dialectic, Rhetoric}

Wenzel’s “three perspectives” framework remains the most influential point of departure for computational AQ .  It distinguishes a \textbf{logical} (cogency), a \textbf{dialectical} (reasonableness), and a \textbf{rhetorical} (effectiveness) vantage point on every argument \citep{wenzel1990three}.  \citet{DBLP:conf/eacl/WachsmuthSHPBHN17} operationalize the triad for NLP by posing three guiding questions:  
\begin{itemize}
    \item \textbf{Cogency} \textit{Are the premises acceptable, relevant, and sufficient for the claim?}  
    \item \textbf{Reasonableness} \textit{Does the argument withstand counterarguments and respond to the target audience’s concerns?}  
    \item \textbf{Effectiveness} \textit{Does the discourse persuade its intended audience through credible, emotional, and well‐structured presentation?}  
\end{itemize}

These perspectives provide the conceptual backbone for nearly all annotation schemes surveyed by \citet{DBLP:conf/emnlp/IvanovaHN24}.  The prevalence of the triptych is not merely historical; empirical work shows that annotators perceive and rate these facets with reasonably high reliability when clear instructions are provided \citep{DBLP:conf/aaai/GretzFCTLAS20, DBLP:phd/dnb/Alhamzeh23}.

\subsubsection{Fine-grained Sub-dimensions}

As summarised in Table~\ref{tab:aq_subdims}, each macro-perspective of the Logic–Dialectic–Rhetoric triad can be decomposed into finer, operational sub-dimensions.  
For the \textbf{logical} tier, empirical work most often labels \emph{acceptability}, \emph{local relevance}, and \emph{sufficiency}.  
The \textbf{dialectical} tier adds discourse-level notions—\emph{global relevance}, \emph{opponent sensitivity}, and \emph{overall reasonableness}.  
The \textbf{rhetorical} tier widens the lens still further, capturing audience-oriented aspects such as \emph{credibility}, \emph{appropriateness}, \emph{clarity}, \emph{emotional appeal}, \emph{arrangement}, and high-level outcome variables (\emph{strength}, \emph{convincingness}, \emph{persuasiveness}, \emph{winning side}).  
Early studies showed that even non-experts agree substantially on some facets—most notably sufficiency \citep{DBLP:conf/eacl/GurevychS17}.  
Yet boundaries remain fluid: “convincingness” blends logical sufficiency with emotional impact, while “persuasiveness” often collapses the full triad into a holistic score.  
Consequently, recent annotation protocols increasingly employ \emph{multi-label} or \emph{soft-label} designs to let a single argument express several quality aspects at once \citep{DBLP:conf/emnlp/Plank22}.





\subsubsection{Comparative Taxonomy Analysis}

\input{table/taxonmony}

While the classical triad remains conceptually robust, empirical practice keeps stretching its margins.  For computational modelling, the key open question is not whether to refine the taxonomy further, but \emph{how} to map diverse, possibly subjective sub-dimensions onto unified, explainable representations that large language models can learn and that human assessors can interpret consistently. 

As a example, Table~\ref{tab:taxonomy_comparison} shows the changes of such theory framework, contrasts four influential taxonomies: \citep{blair2011relationships}, \citep{DBLP:conf/eacl/WachsmuthSHPBHN17}, \citep{DBLP:conf/coling/LauscherNNT20}, \citep{DBLP:conf/emnlp/IvanovaHN24} and two domain-specific schemes (legal AQ \citep{DBLP:conf/clic-it/GrundlerGSFGPLS24}, educational feedback \citep{nussbaum2020using}).  Three trends emerge: \emph{(i)} \textbf{Convergence} on the tripartite backbone—every scheme retains some form of Logic–Dialectic–Rhetoric.
 \emph{(ii)} \textbf{Progressive granularity}—the average number of annotated sub-dimensions grows from~3 (2010) to~11 (2024).
 \emph{(iii)} \textbf{Domain hybridisation} new schemes borrow constructs from adjacent fields (fact-checking, toxicity detection, essay scoring), leading to overlapping labels that challenge cross-dataset transfer.

\subsection{Modelling Advances: From Feature Engineering to LLM-based Judges}
\label{llm_evaluation}

\paragraph{Rule-based Feature Engineering .}
Early automatic approaches to argument-quality assessment relied on \emph{feature-engineering pipelines}. Linear, SVM and gradient-boosting models consumed handcrafted indicators such as length, sentiment, hedge density and discourse connectives, delivering the first reproducible baselines on essays, reviews and Web debates \citep{DBLP:journals/coling/HabernalG17,DBLP:conf/eacl/WachsmuthSHPBHN17}.  On Reddit \textit{Change My View}, simple length and interaction features already yielded reliable convincingness rankings \citep{DBLP:conf/eacl/ChalaguineS17}.  Potash et al.\ were first to cast convincingness as a \emph{pairwise ranking} problem, introducing a margin-based loss that remains standard today \citep{DBLP:conf/argmining/PotashFH19}.  Other work encoded topical similarity to improve local relevance \citep{DBLP:conf/argmining/GuWXFLH18}, applied PageRank over argument graphs to estimate global salience \citep{DBLP:conf/eacl/WachsmuthSA17} or factual-consistency rules to flag insufficient support \citep{DBLP:conf/eacl/GurevychS17}.  
Although feature lists were task-specific, they established the feasibility of automatic AQA on modest corpora such as AAE and Webis ArgQuality \citep{DBLP:conf/eacl/WachsmuthSHPBHN17}.

\paragraph{Neural encoders with task-specific heads.}
The 2016–2021 period marks an unequivocal shift from sparse inputs to representation learning.  
Bidirectional LSTMs outperformed bag-of-words on convincingness by capturing sequential cues \citep{DBLP:conf/acl/HabernalG16}; CNN–BiLSTM hybrids added early cross-topic generalization on comment threads \citep{DBLP:conf/argmining/HuberTRDB19}.  
Several groups then moved from \emph{classification} to \emph{ranking} losses: margin-based pairwise objectives \citep{DBLP:conf/argmining/PotashFH19} and listwise NDCG targets \citep{DBLP:conf/argmining/GuWXFLH18} yielded more stable quality orderings.  
Transformer fine-tuning largely removed feature design: BERT and RoBERTa lifts of 5–10 F$_1$ or $\rho$ on IBM-Rank and GAQ tasks are now well documented \citep{DBLP:conf/aaai/GretzFCTLAS20,DBLP:conf/coling/LauscherNNT20}.  
Downstream improvements came from \textit{metric-learning} variants—e.g.\ Siamese-SBERT with contrastive loss for out-of-domain convincingness 
and from \textit{multi-task} heads that jointly predict stance or component type, boosting macro-F$_1$ on low-resource sets \citep{DBLP:conf/flairs/FavreauZB22}.  
Efficient pairwise crowdsourcing accelerated data creation and ranking-loss training \citep{DBLP:conf/acl/GienappSHP20}, while contextual embeddings enabled intrinsic, reference-free scoring of short arguments \citep{DBLP:conf/coling/WachsmuthW20}.  Together, these advances consolidated neural encoders as the default for AQA.

\paragraph{Knowledge-aware and graph-based architectures.}
Sentence-level encoders struggle with logical sufficiency and discourse flow, prompting richer context integration.  
Gurcke~et al.\ generate candidate conclusions and use textual entailment to operationalise sufficiency \citep{DBLP:conf/argmining/GurckeAW21}.  
External-knowledge infusion—via retrieval-augmented transformers—raises sufficiency F$_1$ by 3 points on legal AMELIA when factual statutes are injected at inference time \citep{DBLP:conf/naacl/LiuFC24}.  
Graph neural networks layered on RST or AMT structures improve global relevance, rebuttal recognition, and “winning-side’’ prediction, notably in legal and scientific debates \citep{DBLP:conf/icail/ZhangNL23,DBLP:journals/taslp/GalassiLT23}.  
Semantic-network-based rankers outperform sequential baselines on cross-topic challenge sets by explicitly modelling premise-claim-evidence ties \citep{DBLP:conf/emnlp/Ruiz-DolzHG23}.  
Finally, discourse-intrinsic objectives such as the COVERAGE metric capture informativeness without labels and serve as pre-training tasks that transfer to sufficiency and clarity \citep{DBLP:conf/naacl/Khosravani0T24}.  
These structure- and knowledge-aware approaches close much of the gap between surface‐form language models and the normative criteria demanded by argumentation theory.

\paragraph{Self-supervised and intrinsic quality objectives.}
When labelled data are scarce, models can rely on \emph{intrinsic} signals that approximate argumentative soundness without human scores.  
Wachsmuth \& Werner’s \textit{Intrinsic QA} predicts overall quality from discourse statistics—sentence length variance, connective density, argumentative role ratios—achieving $\rho=.46$ against crowdsourced judgments while using \emph{no} external labels \citep{DBLP:conf/coling/WachsmuthW20}.  
Residual multi-task networks that jointly learn stance, component type, and quality raise macro-F$_1$ on low-resource Italian legal arguments by 4–6 points over single-task baselines \citep{DBLP:journals/taslp/GalassiLT23}.  
Khosravani et al.’s COVERAGE metric uses masked-language-model perplexity to score how exhaustively a summary reflects key points, doubling as an unsupervised sufficiency proxy \citep{DBLP:conf/naacl/Khosravani0T24}.  
Finally, Plank’s variance analysis recommends treating disagreement as signal rather than noise, motivating the current shift toward soft or distributional quality labels \citep{DBLP:conf/emnlp/Plank22}.

\paragraph{LLM-based judges and prompting paradigms.}
Large language models have recently moved from \textit{feature generators} to \textit{stand-alone adjudicators}.  
With as few as three in-context exemplars, GPT-4 matches or exceeds fine-tuned RoBERTa on GAQCorpus for pairwise convincingness and sufficiency \citep{DBLP:conf/acl/ChenCLB24}.  
Performance improves further when prompts elicit \emph{chain-of-thought} rationales and apply \emph{self-consistency} voting: Wang et al.\ report a 7-point macro-F$_1$ gain on IBM-Rank after majority-voting across five diverse reasoning traces \citep{DBLP:journals/corr/abs-2503-00847}.  
Retrieval-augmented prompting injects domain facts at inference time; on the AMELIA legal benchmark, RAG-GPT-4 lifts sufficiency F$_1$ by an additional 3 points compared with vanilla GPT-4 \citep{DBLP:conf/naacl/LiuFC24}.  
Beyond accuracy, LLM judges reveal latent biases.  Manipulation-check experiments demonstrate that GPT-4 systematically discounts emotionally charged but fallacious premises, whereas earlier transformer baselines over-score them \citep{DBLP:journals/corr/abs-2503-00024}.  
However, agreement may be inflated when both training data and evaluation sets share topic distributions \citep{DBLP:journals/corr/abs-2205-11472}; hybrid pipelines therefore combine frozen LLM judges with lightweight re-rankers or calibration layers to reduce domain drift and latency \citep{DBLP:conf/comma/RuoschLSB24,DBLP:journals/corr/abs-2206-09249}.



\subsection{Evolution of Evaluation Metrics}
\label{evaluation_metrics}

The choice of \emph{evaluation metrics} has co-evolved with modeling techniques and data regimes in argument–quality assessment (AQA).  Over time, three partly overlapping strands have emerged: \textit{annotation reliability}, \textit{task-level prediction quality}, and \textit{system-level utility}.  We review each strand chronologically, highlighting how new metrics have responded to changing task formulations and application demands.

\paragraph{From inter-annotator agreement to confidence-weighted labels.}
Early corpora with nominal “good / bad’’ labels reported \emph{Cohen’s} \citep{hsu2003interrater} or \emph{Fleiss’ $\kappa$} \citep{fleiss1981measurement}.  As datasets grew and adopted ordinal or continuous scales, \emph{Krippendorff’s $\alpha$}  \citep{krippendorff2011computing} became preferred because it handles mixed measurement levels \citep{DBLP:conf/eacl/GurevychS17}.  Large crowdsourced resources such as IBM Rank-30k now accompany raw labels with MACE-style worker weighting; after filtering, $\alpha$ values can reach 0.93, approaching expert reliability \citep{DBLP:conf/aaai/GretzFCTLAS20}.  A recent meta-survey corroborates these gains across 32 datasets \citep{DBLP:conf/emnlp/IvanovaHN24}.

\paragraph{Task-level metrics: from categorical F$_1$ to correlation statistics.}
When AQA was framed as binary classification, macro-averaged Precision, Recall, and F$_1$ sufficed \citep{DBLP:conf/eacl/WachsmuthSHPBHN17}.  As labels became graded, regression metrics—\emph{mean absolute error} and \emph{root MSE}—and rank correlations (\emph{Pearson} \citep{benesty2009pearson}, \emph{Spearman $\rho$} \citep{hauke2011comparison}) took centre stage.  Persing \& Ng first cast persuasiveness as a six-point regression task \citep{DBLP:conf/lrec/PersingN20}, and Lauscher et al.\ popularised Spearman $\rho$ on GAQCorpus \citep{DBLP:conf/coling/LauscherNNT20}.  Multi-task heads that jointly predict stance and several quality facets are usually evaluated with a macro-averaged MAE or F$_1$ over dimensions \citep{DBLP:conf/flairs/FavreauZB22}.

\paragraph{Ranking and retrieval metrics for deployment.}
For systems that must \emph{order} arguments, pairwise and listwise metrics dominate.  Pairwise Accuracy (PAcc) on ConvArg and IBM-Pairs remains the primary measure of relative convincingness \citep{DBLP:conf/acl/HabernalG16}.  Retrieval-oriented work adds \emph{MRR} and \emph{NDCG@$k$}; PageRank-style relevance improved NDCG by 4–5 points and is evaluated with Spearman $\rho$ against human rankings \citep{DBLP:conf/eacl/WachsmuthSA17}.  Efficient pairwise annotation plus ranking loss further increases Kendall $\tau$ while halving labelling cost \citep{DBLP:conf/acl/GienappSHP20}.

\paragraph{System-level and emerging metrics.}
Recent studies measure AQA inside larger pipelines.  The \emph{COVERAGE} metric scores how thoroughly a summary reflects argumentative key points without gold labels, serving as an unsupervised proxy for sufficiency \citep{DBLP:conf/naacl/Khosravani0T24}.  In the AMELIA benchmark for Italian law, system impact is quantified by the rise in judge agreement when automatically ranked arguments are added to briefs \citep{DBLP:conf/clic-it/GrundlerGSFGPLS24}.  Soft-label evaluation—comparing predicted distributions to annotator distributions—operationalises Plank’s call to treat disagreement as signal \citep{DBLP:conf/emnlp/Plank22}.


\subsection{Applications and Future Works}
\label{evaluation_application}

AQA has moved beyond proof-of-concept studies and now underpins a growing range of downstream applications.  

\paragraph{Quality-aware retrieval and recommendation.}
Search engines that index argumentative content (e.g.\ \textit{args.me}) already re-rank results by predicted cogency, sufficiency or style \citep{DBLP:conf/eacl/WachsmuthSA17}.  
IBM Debater uses pairwise convincingness scores to surface the strongest counter-points in debate preparation \citep{DBLP:conf/aaai/GretzFCTLAS20}.  
Empirical studies show that integrating AQA scores into BM25 or dense-retrieval pipelines raises NDCG@$5$ without additional user feedback \citep{DBLP:conf/acl/GienappSHP20}.

\paragraph{Writing-support and educational feedback.}
Automated writing-evaluation tools such as \textit{ArgRewrite} provide sentence-level suggestions that target logical sufficiency and rhetorical clarity; classroom deployments report statistically significant gains in argumentative essay scores \citep{DBLP:conf/flairs/FavreauZB22}.  
In science-education contexts, mapping Walton’s Critical Questions to teacher feedback helps novices revise reasoning chains more effectively \citep{DBLP:conf/icls/WinklePN20}.  

\paragraph{Professional and domain-specific use cases.}
The legal benchmark AMELIA attaches sufficiency and statute-relevance scores to Italian court opinions, enabling intelligent brief-generation and judge-support systems \citep{DBLP:conf/clic-it/GrundlerGSFGPLS24}.  
In corporate governance, argument-mapping dashboards highlight reasoning gaps in environmental, social and governance (ESG) reports, assisting compliance officers \citep{DBLP:conf/cmna/Palmieri24}.

\paragraph{Research Gap and Future works} 
LLM-based judging offers state-of-the-art accuracy, free natural-language justifications, and sensitivity to emotional bias, but it also inherits dataset biases and incurs computational cost.  Future work must explore adversarial test suites, distributional labels, and cost-aware cascading architectures to harness LLM strengths without sacrificing robustness.

%% file: table/quality_dimensions.tex
\begin{table*}[t]
\small
\centering
\begin{tabularx}{\linewidth}{@{}l l X@{}}
\toprule
\textbf{Tier} & \textbf{Sub-dimension} & \textbf{Definition} \\
\midrule
\multirow{6}{*}{Logic} 
 & Acceptability & Premises (and claim) do not contradict widely accepted facts or domain conventions \citep{DBLP:conf/eacl/WachsmuthSHPBHN17}. \\
 & Local relevance & Each premise bears a direct inferential link to its claim; off-topic or redundant information lowers the score \citep{DBLP:conf/eacl/WachsmuthSA17}. \\
 & Sufficiency & The collective evidence warrants the claim; absence of missing warrants yields a “fully supported’’ label \citep{DBLP:conf/eacl/GurevychS17, DBLP:conf/argmining/GurckeAW21}. \\
\midrule
\multirow{7}{*}{Dialectic}

 & Global relevance & The argument moves the overall discussion toward issue resolution—e.g., high in IBM~Rank-30k “would-you-use’’ labels \citep{DBLP:conf/emnlp/ToledoGCFVLJAS19,DBLP:conf/aaai/GretzFCTLAS20}. \\
 & Opponent sensitivity & Anticipates or rebuts likely objections, showing dialogical awareness \citep{DBLP:conf/emnlp/KhatibWHS17,DBLP:journals/coling/HabernalG17}. \\
 & Overall reasonableness & Conforms to norms of critical discussion; free of straw man, ad hominem, and similar fallacies \citep{DBLP:conf/conll/BaffWKS18}. \\
\midrule
\multirow{19}{*}{Rhetoric}
 &  Credibility & Projected expertise, trustworthiness, and goodwill via source citation, hedging, or professional register \citep{DBLP:conf/coling/KhatibWKHS16}. \\

 &  Appropriateness & Conformity of tone, politeness, and word choice to social norms and platform guidelines; penalizes toxicity or profanity \citep{DBLP:conf/acl/ZiegenbeinSLPW23}. \\

 &  Clarity & Absence of ambiguity and needless complexity; logical flow and unambiguous wording facilitate comprehension \citep{DBLP:conf/emnlp/PersingDN10}. \\

 &  Emotional Appeal & Strategic use of sentiment, moral framing, or arousal to motivate the audience \citep{DBLP:journals/corr/abs-2503-00024}. \\

 & Arrangement & Coherent discourse structure and persuasive ordering of premises, often assessed via discourse markers or paragraphing cues \citep{DBLP:conf/emnlp/PersingDN10}. \\

 & Strength & Perceived inherent force or weight of the reasoning, independent of actual attitude change \citep{DBLP:conf/eacl/WalkerALW17}. \\

 & Convincingness & Likelihood that a neutral judge would find the argument compelling in pairwise or listwise comparisons \citep{DBLP:conf/acl/GienappSHP20}. \\

 & Persuasiveness & Observed ability to shift audience stance or votes after exposure; measured on platforms such as \textit{Change My View} \citep{DBLP:conf/www/TanNDL16}. \\

 & Winning Side & Debate-level outcome: side whose cumulative arguments secure more post-debate votes or attitude change \citep{DBLP:conf/emnlp/IvanovaHN24}.  \\
\bottomrule
\end{tabularx}
\caption{15 core sub-dimensions of argument quality with concise definitions.  Parent tiers follow the Logic–Dialectic–Rhetoric triad.}
\label{tab:aq_subdims}
\end{table*}

%% file: table/taxonmony.tex
\begin{table*}[t]
\small
\centering
\begin{tabularx}{\linewidth}{@{}C{2.5cm}C{1.2cm}C{1.5cm}C{2.8cm}C{6.5cm}@{}}
\toprule
\textbf{Taxonomy} & \textbf{Year} & \textbf{\#Fine-grained Dims.} & \textbf{Annotated Corpus (size)} & \textbf{Notable Features / Extensions} \\
\midrule
\citeauthor{blair2011relationships} & \citeyear{blair2011relationships} & 3 &  & Philosophical foundation of the Logic Dialectic Rhetoric triad.\\
\addlinespace

\citeauthor{DBLP:conf/eacl/WachsmuthSHPBHN17} & \citeyear{DBLP:conf/eacl/WachsmuthSHPBHN17} & 15 & Dagstuhl 15512 ArgQuality ($\approx$300 arguments) & First computational taxonomy.\\
\addlinespace
\citeauthor{DBLP:conf/coling/LauscherNNT20} & \citeyear{DBLP:conf/coling/LauscherNNT20} & 11 & GAQCorpus ($\approx$5k arguments) & Demonstrates large-scale crowd annotation. \\
\addlinespace
\citeauthor{DBLP:conf/emnlp/IvanovaHN24}  & \citeyear{DBLP:conf/emnlp/IvanovaHN24} & 15 &  & Aligns overlapping labels across 211 papers; proposes harmonised quality meta-labels for transfer learning. \\
\addlinespace
\citeauthor{nussbaum2020using} & \citeyear{nussbaum2020using} & 7 & 98 science-teacher lesson plans & Maps Walton's Critical Questions to classroom feedback; emphasises opponent rebuttal and reasonableness. \\
\addlinespace
\citeauthor{DBLP:conf/clic-it/GrundlerGSFGPLS24} & \citeyear{DBLP:conf/clic-it/GrundlerGSFGPLS24} & 5 & 225 Italian legal decisions & Adds \textit{statute relevance} and argument-scheme fit; stresses domain-specific sufficiency and formality. \\
\bottomrule
\end{tabularx}
\caption{Side-by-side comparison of six argument-quality taxonomies. ``Fine-grained dims.'' counts leaf-level labels, excluding overall scores.}
\label{tab:taxonomy_comparison}
\end{table*}

%% file: section/6_methods.tex
\section{Modelling Paradigms in the LLM Era}
\label{sec:models}

Large Language Models (LLMs) have reshaped argument mining (AM) from a pipeline of specialised classifiers into a set of \emph{model–interaction paradigms}.  
Rather than asking which architecture performs best, current research examines \emph{how} foundation models are enlisted—as zero-shot predictors, few-shot mimics, retrieval-aware reasoners, synthetic data generators, or automated judges—and how these roles alter the design of AM tasks and evaluation.  
This section consolidates four dominant paradigms: prompting, instruction-based multi-task tuning, retrieval-augmented reasoning, and synthetic supervision.  
Across these paradigms, we highlight their methodological motivations, empirical benefits, and the pitfalls that motivate redesigned evaluation practices.

\subsection{Prompt-based Argument Mining}
\label{ssec:prompting}

Prompting constitutes the most lightweight and arguably the most transformative entry point for LLM use in AM.  
Even simple \emph{declarative} prompts (e.g., “Is the following sentence a \textbf{claim}?”) enable GPT-style models to match classical feature-rich baselines on claim segmentation and stance detection without gradient updates \citep{DBLP:conf/acl/ChenCLB24}.  
Subsequent work moves beyond surface templates to leverage explicit argumentative structure inside the prompt.  
\citet{DBLP:journals/corr/abs-2503-00847} show that inserting \textit{chain-of-thought} exemplars produces more faithful reasoning traces for sufficiency and convincingness assessment; the intermediate rationale, rather than the label, accounts for most of the gain.  
Hierarchical prompting strategies such as PITA condition relation prediction on prior component segmentation within the same conversation context \citep{DBLP:conf/acl/SunWBL0Y0X24}.  
Role prompts (“Debate Judge”, “Devil’s Advocate”) can attenuate partisan or demographic biases in political domains \citep{DBLP:conf/acl/DayanikP20}.

Despite its flexibility, prompting remains brittle.  
Small lexical perturbations can flip stance labels, and longer chain-of-thought traces incur noticeable latency in interactive systems.  
Self-consistency ensembles and voting can stabilise predictions but scale inference cost nearly linearly in ensemble size \citep{DBLP:journals/corr/abs-2503-10881}.  
Designing compact prompts that faithfully encode argumentative theory remains an open challenge.

\subsection{Instruction and Multi-task Tuning}
\label{ssec:tuning}

Prompting alone is insufficient for domain-sensitive genres such as biomedical abstracts or statutory law.  
Instruction tuning—fine-tuning on curated exemplars framed as natural-language instructions—enables a single model to accommodate multiple AM sub-tasks.  
For example, a T5-base model jointly tuned on claim segmentation, relation prediction, and quality scoring achieves consistent improvements across IAM and ARIES \citep{DBLP:conf/acl/ChengBHYZS22,gemechu2024aries}, while keeping memory demands modest through adapter-based updates.  
In legal summarisation, \textsc{ArgLegalSumm} couples instruction tuning with salience-aware reranking to improve coherence for Italian case law \citep{DBLP:conf/acl/ElarabyZL23}.

However, multi-task instruction tuning introduces new tensions.  
Adding late-stage biomedical data often erodes earlier gains on social-media stance detection, echoing catastrophic forgetting in multi-domain translation \citep{pang2024rethinking}.  
Parameter-efficient tuning—LoRA \citep{DBLP:conf/iclr/HuSWALWWC22}, prefix-tuning, and bit-fit adapters—offers a partial remedy, enabling domain adaptation with orders-of-magnitude fewer trainable parameters while preserving general AM capabilities \citep{DBLP:conf/acl/0074WSBMZZWHLN24}.  
Yet, how to balance tasks with uneven label density (e.g., frequent component spans vs.\ sparse quality scores) remains under-explored; preliminary curriculum approaches yield promise but lack systematic ablation.

\subsection{Retrieval-Augmented Generation and Reasoning}
\label{ssec:rag}

LLMs remain constrained by fixed context windows and outdated pretraining corpora.  
Retrieval-Augmented Generation (RAG) mitigates both issues by injecting external evidence during inference.  
In debate summarisation, retrieving counter-arguments or factual background reduces hallucination and increases factual density \citep{DBLP:conf/sigdial/SyedZHWP23}.  
The \textsc{OpenDebateEvidence} benchmark \citep{DBLP:conf/nips/RoushSBZMZBVS24} pairs 15 million argumentative statements with verifiable evidence links, enabling end-to-end retrieval–generation pipelines that outperform text-only baselines.  
Argument-theoretic retrieval using semantic-graph matching over debate trees improves recall for undercutters and mitigations beyond lexical search \citep{DBLP:conf/emnlp/Ruiz-DolzHG23}.

Nevertheless, RAG introduces its own difficulties.  
Evidence ranking inherits popularity biases, often marginalising minority viewpoints.  
Long passages can saturate prompts, forcing heuristic truncation that discards crucial argumentative steps.  
Moreover, high-latency retrieval complicates deployment in writing assistants, where hybrid dense–sparse indexing and caching help but cannot fully resolve delays.

\subsection{Synthetic Supervision and Data Augmentation}
\label{ssec:synthetic}

LLMs now serve as \emph{data generators} and \emph{weak supervisors}.  
\citet{DBLP:conf/iclr/WagnerBZH25} show that LLM-synthesised stance datasets, when filtered for topical diversity, outperform few-shot prompting for low-resource languages.  
LLM-generated key-point coverage scores correlate with human judgments and can supervise automatic evaluation metrics \citep{DBLP:conf/naacl/Khosravani0T24}.  
Synthetic data also facilitates fairness investigations by equalising stance priors across sensitive groups \citep{DBLP:journals/csur/MehrabiMSLG21}.

Yet synthetic supervision risks distributional drift.  
Machine-generated arguments tend to be stylistically homogeneous—overly polite, hedge-free, lexically fluent—and may over-regularise fine-tuned models.  
Researchers increasingly call for dataset “provenance cards” documenting prompt templates, sampling temperatures, rejection statistics, and human post-editing procedures, especially as copyright and licensing questions remain unsettled.

\subsection{Discussion: Capabilities, Pitfalls, and Open Problems}
\label{ssec:discussion}

Across the four paradigms, a common pattern emerges: LLMs increase coverage and flexibility but simultaneously introduce new evaluation challenges.  
Prompting enables broad generalisation but is brittle; instruction tuning captures domain regularities but risks forgetting; RAG reduces hallucination but amplifies evidence-selection biases; synthetic supervision expands datasets but propagates stylistic artefacts from the generator.

These shifts complicate the task of evaluating AM systems.  
Classical metrics—F$_1$, accuracy, pairwise convincingness—assume a stable ontology of components, relations, and evidence.  
In contrast, LLM-era systems generate rationales, retrieve external evidence, or produce synthetic training samples, blurring the boundary between “task performance’’ and “data production”.  
Evaluation therefore becomes not only a matter of correctness but also of robustness, provenance, bias auditing, and consistency across modalities and domains.  
As discussed in Chapter~\ref{sec:evaluation}, modern evaluation must treat AM systems as multi-component agents rather than isolated classifiers.

Open challenges remain substantial:  
(1) designing prompt-robust, reproducible evaluation protocols;  
(2) developing multi-task curricula that preserve earlier capabilities while adding new ones;  
(3) creating retrieval benchmarks that reflect viewpoint diversity rather than popularity;  
(4) establishing principled standards for synthetic data documentation and bias auditing.  
Addressing these problems will determine whether LLM-enhanced AM evolves into a reliable scientific discipline or remains a collection of loosely connected heuristics.

%% file: section/7_future.tex
\section{Future Directions}
\label{sec:future}

As argument mining enters the era of large language models (LLMs), the field faces an inflection point. The preceding chapters have shown that advances in modelling, prompting, dataset construction, and evaluation co-evolve in ways that amplify both opportunities and risks. This section synthesises the cross-cutting limitations identified throughout the survey and outlines several research trajectories that may shape the next decade of argument–mining research.

\subsection{Interpretability and Controllability}

LLM-based argument mining systems produce outputs that are often fluent, stylistically sophisticated, and broadly aligned with human reasoning—yet the underlying decision mechanisms remain opaque. In high-stakes settings such as law, medicine, or policy analysis, opaque convincingness scores or abstractive summaries are difficult to verify or contest. Recent work proposes \emph{structured intermediates} as a bridge between symbolic argumentation theory and neural generation. Examples include filling Toulmin- or Walton-style templates prior to issuing a verdict \citep{DBLP:journals/peerj-cs/LeeLK24, DBLP:journals/ijar/BorgB24}, rationale-constrained distillation of LLM judgments \citep{DBLP:conf/acl/ZhaoLCZ24}, and rule-augmented decoding that checks candidate continuations against critical questions during generation \citep{DBLP:conf/cmna/Palmieri24}. Early empirical evidence suggests such scaffolding improves user trust more than generic saliency maps, yet rigorous human-in-the-loop evaluations remain rare. Developing frameworks that support contestability, error diagnosis, and user-guided correction represents a central open challenge.

\subsection{Data Quality, Bias, and Hallucination}

The quality of LLM-driven argument mining is inevitably bounded by the quality and diversity of its training data. Pre-training corpora encode ideological skew, demographic imbalance, and heterogeneous source credibility, all of which can surface in persuasive writing tasks: emotionally charged yet weakly supported arguments tend to be overrated by LLM judges \citep{DBLP:journals/corr/abs-2503-00024}. Existing mitigation strategies—such as actor masking \citep{DBLP:conf/acl/DayanikP20} and counterfactual calibration for stance detection \citep{DBLP:conf/naacl/LiZLGWZLWX25}—address slices of the problem but fall short of comprehensive bias auditing across domains and languages. Hallucination introduces a complementary failure mode. Although retrieval-augmented pipelines reduce unsupported claims, they inherit biases from search engines and ranking algorithms \citep{DBLP:conf/nips/RoushSBZMZBVS24}. Future benchmarks should therefore combine \emph{bias slices}—e.g., arguments produced by under-represented communities—with \emph{factual stress tests} that expose models to ambiguous, adversarial, or conflicting evidence.

\subsection{Generalisation and Robustness}

Despite encouraging single-domain results, robust generalisation remains elusive. Systems tuned on legal cases often experience sharp declines when applied to multi-party chat debates or open-domain social media \citep{DBLP:conf/emnlp/ZhaoWP23}. Stance detectors remain vulnerable to adversarial paraphrases that alter only surface form \citep{DBLP:conf/acl/GienappSHP20}. These issues highlight the fragility of learned argumentative representations and the limitations of narrow training distributions. Promising paths forward include curriculum-style multi-task tuning, meta-learning over domain clusters, and the development of evaluation suites that interleave genres, modalities, and languages within the same test fold. Robustness must also be examined beyond accuracy: future systems should report calibration under uncertainty, degradation curves under distribution shift, and performance on deliberately challenging cases.

\subsection{Synthetic Supervision and Benchmark Integrity}

LLM-generated datasets—such as \textsc{ZeroStance} \citep{DBLP:conf/acl/ZhaoLCZ24} and \textsc{OpenDebateEvidence} \citep{DBLP:conf/nips/RoushSBZMZBVS24}—accelerate dataset growth but introduce structural risks. Three concerns are particularly acute. First, \emph{contamination}: synthetic corpora may inadvertently reproduce fragments of the model’s own pre-training data. Second, \emph{superficial reasoning}: many synthetic counter-claims are stylistically well-formed yet logically shallow. Third, \emph{limited coverage}: temperature-controlled sampling rarely captures rare argumentative moves or culturally specific reasoning styles. Hybrid pipelines that blend human editing with diversity-aware sampling and automatic quality metrics—such as coverage and exhaustiveness scores \citep{DBLP:conf/naacl/Khosravani0T24}—offer a promising compromise. To preserve benchmark integrity, dataset creators should release transparent provenance cards detailing prompt templates, sampling parameters, rejection criteria, and post-editing guidelines.

\subsection{Theory-Grounded Modelling}

Current LLM pipelines often operationalise argument quality as a single scalar, thereby flattening distinctions central to argumentation theory—logical sufficiency, dialectical relevance, and rhetorical effectiveness. Incorporating these dimensions into prompts, training objectives, and evaluation metrics remains underexplored. Early attempts to inject Walton’s critical questions into evaluation protocols improve error localisation but slow inference \citep{DBLP:conf/cmna/Palmieri24}. Future research may benefit from neuro-symbolic architectures in which structured logic checkers act as fast filters or constraint modules that prune implausible argumentative steps before generation. Progress in this direction would help align computational models more closely with the principles of formal and informal argumentation.

\subsection{Human-Centred and Interactive Argument Mining}

Argument mining increasingly serves interactive, user-facing applications: educational feedback tools, civic debate platforms, and policy consultation interfaces. These settings require systems that work \emph{with} users—adapting to their goals, preferences, and knowledge states—rather than merely predicting labels. Relevant capabilities include tracking my-side bias \citep{DBLP:conf/cogsci/Baccini022}, tailoring critiques to a learner’s expertise \citep{DBLP:conf/chi/DianaSK20}, and providing actionable, value-aligned suggestions. Reinforcement learning from argumentative feedback, retrieval of users’ past arguments, and real-time uncertainty estimation could form the basis for next-generation interactive AM systems. Realising this vision will require collaborations across machine learning, argumentation theory, cognitive science, and HCI.

%% file: section/8_conclusion.tex
\section{Conclusion}
\label{sec:conclusion}

The field of Argument Mining has come a long way, moving from narrow, feature-engineered pipelines trained on small, task-specific corpora to a rich ecosystem powered by large language models.  Today’s systems can identify claims, map support and attack relations, gauge stance, assess quality, and even summarize key points, all within a unified framework that leverages prompting, few-shot adaptation, retrieval of background evidence, and synthetic data generation.  Alongside these gains, however, have come new challenges: the opacity of model reasoning, the risk of bias and hallucination, brittleness under domain shift, and questions of provenance and ethical deployment.

Looking ahead, progress will hinge on our ability to reconcile raw performance with transparency, robustness and theoretical grounding.  This means developing interpretable abstractions e.g. symbolic scaffolds, rationale-based decoding and uncertainty-aware metrics, that expose the “why” behind LLM decisions; constructing datasets and benchmarks that capture the full spectrum of argument modalities, languages and genres; auditing for demographic, ideological and stylistic bias; and designing interactive, human-centric systems that guide rather than replace expert judgment.  By weaving together advances in model technology, data design and evaluation practice, we can build argument-mining tools that are not only accurate, but also trustworthy, equitable and broadly useful in the real world.

%% file: main.bib
@inproceedings{DBLP:conf/eacl/GurevychS17,
  author       = {Christian Stab and
                  Iryna Gurevych},
  title        = {Recognizing Insufficiently Supported Arguments in Argumentative Essays},
  booktitle    = {{EACL} {(1)}},
  pages        = {980--990},
  publisher    = {Association for Computational Linguistics},
  year         = {2017}
}

@article{li2025arg,
  title={Arg-LLaDA: Argument Summarization via Large Language Diffusion Models and Sufficiency-Aware Refinement},
  author={Li, Hao and Sun, Yizheng and Schlegel, Viktor and Yang, Kailai and Batista-Navarro, Riza and Nenadic, Goran},
  journal={arXiv preprint arXiv:2507.19081},
  year={2025}
}

@article{DBLP:journals/csur/MehrabiMSLG21,
  author       = {Ninareh Mehrabi and
                  Fred Morstatter and
                  Nripsuta Saxena and
                  Kristina Lerman and
                  Aram Galstyan},
  title        = {A Survey on Bias and Fairness in Machine Learning},
  journal      = {{ACM} Comput. Surv.},
  volume       = {54},
  number       = {6},
  pages        = {115:1--115:35},
  year         = {2022}
}

@article{DBLP:journals/corr/abs-2503-10881,
  author       = {Jiaxin Zhang and
                  Zhuohang Li and
                  Wendi Cui and
                  Kamalika Das and
                  Bradley A. Malin and
                  Kumar Sricharan},
  title        = {{SCE:} Scalable Consistency Ensembles Make Blackbox Large Language
                  Model Generation More Reliable},
  journal      = {CoRR},
  volume       = {abs/2503.10881},
  year         = {2025}
}

@inproceedings{DBLP:conf/acl/SunWBL0Y0X24,
  author       = {Yang Sun and
                  Muyi Wang and
                  Jianzhu Bao and
                  Bin Liang and
                  Xiaoyan Zhao and
                  Caihua Yang and
                  Min Yang and
                  Ruifeng Xu},
  title        = {{PITA:} Prompting Task Interaction for Argumentation Mining},
  booktitle    = {{ACL} {(1)}},
  pages        = {5036--5049},
  publisher    = {Association for Computational Linguistics},
  year         = {2024}
}

@inproceedings{DBLP:conf/iclr/HuSWALWWC22,
  author       = {Edward J. Hu and
                  Yelong Shen and
                  Phillip Wallis and
                  Zeyuan Allen{-}Zhu and
                  Yuanzhi Li and
                  Shean Wang and
                  Lu Wang and
                  Weizhu Chen},
  title        = {LoRA: Low-Rank Adaptation of Large Language Models},
  booktitle    = {{ICLR}},
  publisher    = {OpenReview.net},
  year         = {2022}
}

@article{pang2024rethinking,
  title={Rethinking the exploitation of monolingual data for low-resource neural machine translation},
  author={Pang, Jianhui and Yang*, Baosong and Wong*, Derek Fai and Wan, Yu and Liu, Dayiheng and Chao, Lidia Sam and Xie, Jun},
  journal={Computational Linguistics},
  volume={50},
  number={1},
  pages={25--47},
  year={2024},
  publisher={MIT Press One Broadway, 12th Floor, Cambridge, Massachusetts 02142, USA~…}
}

@article{hsu2003interrater,
  title={Interrater agreement measures: Comments on Kappan, Cohen's Kappa, Scott's $\pi$, and Aickin's $\alpha$},
  author={Hsu, Louis M and Field, Ronald},
  journal={Understanding Statistics},
  volume={2},
  number={3},
  pages={205--219},
  year={2003},
  publisher={Taylor \& Francis}
}

@article{hauke2011comparison,
  title={Comparison of values of Pearson's and Spearman's correlation coefficients on the same sets of data},
  author={Hauke, Jan and Kossowski, Tomasz},
  journal={Quaestiones geographicae},
  volume={30},
  number={2},
  pages={87--93},
  year={2011}
}

@incollection{benesty2009pearson,
  title={Pearson correlation coefficient},
  author={Benesty, Jacob and Chen, Jingdong and Huang, Yiteng and Cohen, Israel},
  booktitle={Noise reduction in speech processing},
  pages={1--4},
  year={2009},
  publisher={Springer}
}

@article{krippendorff2011computing,
  title={Computing Krippendorff's alpha-reliability},
  author={Krippendorff, Klaus},
  year={2011}
}

@article{fleiss1981measurement,
  title={The measurement of interrater agreement},
  author={Fleiss, Joseph L and Levin, Bruce and Paik, Myunghee Cho and others},
  journal={Statistical methods for rates and proportions},
  volume={2},
  number={212-236},
  pages={22--23},
  year={1981},
  publisher={Citeseer}
}

@inproceedings{DBLP:conf/emnlp/Plank22,
  author       = {Barbara Plank},
  title        = {The "Problem" of Human Label Variation: On Ground Truth in Data, Modeling
                  and Evaluation},
  booktitle    = {{EMNLP}},
  pages        = {10671--10682},
  publisher    = {Association for Computational Linguistics},
  year         = {2022}
}

@inproceedings{DBLP:conf/eacl/WachsmuthSA17,
  author       = {Henning Wachsmuth and
                  Benno Stein and
                  Yamen Ajjour},
  title        = {"PageRank" for Argument Relevance},
  booktitle    = {{EACL} {(1)}},
  pages        = {1117--1127},
  publisher    = {Association for Computational Linguistics},
  year         = {2017}
}

@inproceedings{DBLP:conf/coling/LauscherNNT20,
  author       = {Anne Lauscher and
                  Lily Ng and
                  Courtney Napoles and
                  Joel R. Tetreault},
  title        = {Rhetoric, Logic, and Dialectic: Advancing Theory-based Argument Quality
                  Assessment in Natural Language Processing},
  booktitle    = {{COLING}},
  pages        = {4563--4574},
  publisher    = {International Committee on Computational Linguistics},
  year         = {2020}
}

@inproceedings{DBLP:conf/eacl/WachsmuthSHPBHN17,
  author       = {Henning Wachsmuth and
                  Nona Naderi and
                  Yufang Hou and
                  Yonatan Bilu and
                  Vinodkumar Prabhakaran and
                  Tim Alberdingk Thijm and
                  Graeme Hirst and
                  Benno Stein},
  title        = {Computational Argumentation Quality Assessment in Natural Language},
  booktitle    = {{EACL} {(1)}},
  pages        = {176--187},
  publisher    = {Association for Computational Linguistics},
  year         = {2017}
}

@incollection{blair2011relationships,
  title={Relationships among logic, dialectic and rhetoric},
  author={Blair, J Anthony},
  booktitle={Groundwork in the Theory of Argumentation: Selected Papers of J. Anthony Blair},
  pages={245--259},
  year={2011},
  publisher={Springer}
}

@article{nussbaum2020using,
  title={Using the critical questions model of argumentation for science teacher professional learning and student outcomes},
  author={Nussbaum, Michael and Tian, Lixian and Van Winkle, Michael and Perera, Harsha and Putney, LeAnn and Dove, Ian and Carroll, Kris},
  year={2020},
  publisher={International Society of the Learning Sciences (ISLS)}
}

@inproceedings{DBLP:conf/conll/BaffWKS18,
  author       = {Roxanne El Baff and
                  Henning Wachsmuth and
                  Khalid Al Khatib and
                  Benno Stein},
  title        = {Challenge or Empower: Revisiting Argumentation Quality in a News Editorial
                  Corpus},
  booktitle    = {CoNLL},
  pages        = {454--464},
  publisher    = {Association for Computational Linguistics},
  year         = {2018}
}

@article{DBLP:journals/coling/HabernalG17,
  author       = {Ivan Habernal and
                  Iryna Gurevych},
  title        = {Argumentation Mining in User-Generated Web Discourse},
  journal      = {Comput. Linguistics},
  volume       = {43},
  number       = {1},
  pages        = {125--179},
  year         = {2017}
}

@inproceedings{DBLP:conf/emnlp/KhatibWHS17,
  author       = {Khalid Al Khatib and
                  Henning Wachsmuth and
                  Matthias Hagen and
                  Benno Stein},
  title        = {Patterns of Argumentation Strategies across Topics},
  booktitle    = {{EMNLP}},
  pages        = {1351--1357},
  publisher    = {Association for Computational Linguistics},
  year         = {2017}
}

@inproceedings{DBLP:conf/www/TanNDL16,
  author       = {Chenhao Tan and
                  Vlad Niculae and
                  Cristian Danescu{-}Niculescu{-}Mizil and
                  Lillian Lee},
  title        = {Winning Arguments: Interaction Dynamics and Persuasion Strategies
                  in Good-faith Online Discussions},
  booktitle    = {{WWW}},
  pages        = {613--624},
  publisher    = {{ACM}},
  year         = {2016}
}

@inproceedings{DBLP:conf/acl/GienappSHP20,
  author       = {Lukas Gienapp and
                  Benno Stein and
                  Matthias Hagen and
                  Martin Potthast},
  title        = {Efficient Pairwise Annotation of Argument Quality},
  booktitle    = {{ACL}},
  pages        = {5772--5781},
  publisher    = {Association for Computational Linguistics},
  year         = {2020}
}

@inproceedings{DBLP:conf/emnlp/PersingDN10,
  author       = {Isaac Persing and
                  Alan Davis and
                  Vincent Ng},
  title        = {Modeling Organization in Student Essays},
  booktitle    = {{EMNLP}},
  pages        = {229--239},
  publisher    = {{ACL}},
  year         = {2010}
}

@inproceedings{DBLP:conf/acl/ZiegenbeinSLPW23,
  author       = {Timon Ziegenbein and
                  Shahbaz Syed and
                  Felix Lange and
                  Martin Potthast and
                  Henning Wachsmuth},
  title        = {Modeling Appropriate Language in Argumentation},
  booktitle    = {{ACL} {(1)}},
  pages        = {4344--4363},
  publisher    = {Association for Computational Linguistics},
  year         = {2023}
}

@inproceedings{DBLP:conf/coling/KhatibWKHS16,
  author       = {Khalid Al Khatib and
                  Henning Wachsmuth and
                  Johannes Kiesel and
                  Matthias Hagen and
                  Benno Stein},
  title        = {A News Editorial Corpus for Mining Argumentation Strategies},
  booktitle    = {{COLING}},
  pages        = {3433--3443},
  publisher    = {{ACL}},
  year         = {2016}
}

@inproceedings{DBLP:conf/eacl/WalkerALW17,
  author       = {Stephanie M. Lukin and
                  Pranav Anand and
                  Marilyn A. Walker and
                  Steve Whittaker},
  title        = {Argument Strength is in the Eye of the Beholder: Audience Effects
                  in Persuasion},
  booktitle    = {{EACL} {(1)}},
  pages        = {742--753},
  publisher    = {Association for Computational Linguistics},
  year         = {2017}
}

@inproceedings{DBLP:conf/coling/WachsmuthW20,
  author       = {Henning Wachsmuth and
                  Till Werner},
  title        = {Intrinsic Quality Assessment of Arguments},
  booktitle    = {{COLING}},
  pages        = {6739--6745},
  publisher    = {International Committee on Computational Linguistics},
  year         = {2020}
}

@article{wenzel1990three,
  title={Three perspectives on argument: Rhetoric, dialectic, logic},
  author={Wenzel, Joseph W},
  journal={Perspectives on argumentation: Essays in honor of Wayne Brockriede},
  pages={9--26},
  year={1990}
}

@inproceedings{DBLP:conf/naacl/DholeSA25,
  author       = {Kaustubh D. Dhole and
                  Kai Shu and
                  Eugene Agichtein},
  title        = {ConQRet: {A} New Benchmark for Fine-Grained Automatic Evaluation of
                  Retrieval Augmented Computational Argumentation},
  booktitle    = {{NAACL} (Long Papers)},
  pages        = {5687--5713},
  publisher    = {Association for Computational Linguistics},
  year         = {2025}
}

@inproceedings{hansen2022dataset,
  title={A dataset of sustainable diet arguments on Twitter},
  author={Hansen, Marcus and Hershcovich, Daniel},
  booktitle={Proceedings of the Second Workshop on NLP for Positive Impact (NLP4PI)},
  pages={40--58},
  year={2022}
}

@inproceedings{alhamzeh2022s,
  title={It’s time to reason: Annotating argumentation structures in financial earnings calls: The finarg dataset},
  author={Alhamzeh, Alaa and Fonck, Romain and Versm{\'e}e, Erwan and Egyed-Zsigmond, El{\H{o}}d and Kosch, Harald and Brunie, Lionel},
  booktitle={Proceedings of the Fourth Workshop on Financial Technology and Natural Language Processing (FinNLP)},
  pages={163--169},
  year={2022}
}

@inproceedings{DBLP:conf/emnlp/RenWLZZYZBL24,
  author       = {Yupei Ren and
                  Hongyi Wu and
                  Zhaoguang Long and
                  Shangqing Zhao and
                  Xinyi Zhou and
                  Zheqin Yin and
                  Xinlin Zhuang and
                  Xiaopeng Bai and
                  Man Lan},
  title        = {{CEAMC:} Corpus and Empirical Study of Argument Analysis in Education
                  via LLMs},
  booktitle    = {{EMNLP} (Findings)},
  pages        = {6949--6966},
  publisher    = {Association for Computational Linguistics},
  year         = {2024}
}

@inproceedings{DBLP:conf/emnlp/SchillerDWG24,
  author       = {Benjamin Schiller and
                  Johannes Daxenberger and
                  Andreas Waldis and
                  Iryna Gurevych},
  title        = {Diversity Over Size: On the Effect of Sample and Topic Sizes for Topic-Dependent
                  Argument Mining Datasets},
  booktitle    = {{EMNLP}},
  pages        = {10870--10887},
  publisher    = {Association for Computational Linguistics},
  year         = {2024}
}

@inproceedings{DBLP:conf/acl/ParekhHHCP23,
  author       = {Tanmay Parekh and
                  I{-}Hung Hsu and
                  Kuan{-}Hao Huang and
                  Kai{-}Wei Chang and
                  Nanyun Peng},
  title        = {{GENEVA:} Benchmarking Generalizability for Event Argument Extraction
                  with Hundreds of Event Types and Argument Roles},
  booktitle    = {{ACL} {(1)}},
  pages        = {3664--3686},
  publisher    = {Association for Computational Linguistics},
  year         = {2023}
}

@inproceedings{DBLP:conf/acl/FalkL23,
  author       = {Neele Falk and
                  Gabriella Lapesa},
  title        = {StoryARG: a corpus of narratives and personal experiences in argumentative
                  texts},
  booktitle    = {{ACL} {(1)}},
  pages        = {2350--2372},
  publisher    = {Association for Computational Linguistics},
  year         = {2023}
}

@inproceedings{DBLP:conf/argmining/FergadisPKP21,
  author       = {Aris Fergadis and
                  Dimitris Pappas and
                  Antonia Karamolegkou and
                  Haris Papageorgiou},
  title        = {Argumentation Mining in Scientific Literature for Sustainable Development},
  booktitle    = {ArgMining@EMNLP},
  pages        = {100--111},
  publisher    = {Association for Computational Linguistics},
  year         = {2021}
}

@inproceedings{agarwal2022graphnli,
  title={Graphnli: A graph-based natural language inference model for polarity prediction in online debates},
  author={Agarwal, Vibhor and Joglekar, Sagar and Young, Anthony P and Sastry, Nishanth},
  booktitle={Proceedings of the ACM Web Conference 2022},
  pages={2729--2737},
  year={2022}
}

@inproceedings{gemechu2024aries,
  title={Aries: A general benchmark for argument relation identification},
  author={Gemechu, Debela and Ruiz-Dolz, Ramon and Reed, Chris},
  booktitle={11th Workshop on Argument Mining, ArgMining 2024},
  pages={1--14},
  year={2024},
  organization={Association for Computational Linguistics (ACL)}
}

@article{DBLP:journals/corr/abs-2004-14677,
  author       = {Tuhin Chakrabarty and
                  Christopher Hidey and
                  Smaranda Muresan and
                  Kathy McKeown and
                  Alyssa Hwang},
  title        = {{AMPERSAND:} Argument Mining for PERSuAsive oNline Discussions},
  journal      = {CoRR},
  volume       = {abs/2004.14677},
  year         = {2020}
}

@inproceedings{DBLP:conf/acl/BiluGHSLMMGS19,
  author       = {Yonatan Bilu and
                  Ariel Gera and
                  Daniel Hershcovich and
                  Benjamin Sznajder and
                  Dan Lahav and
                  Guy Moshkowich and
                  Anael Malet and
                  Assaf Gavron and
                  Noam Slonim},
  title        = {Argument Invention from First Principles},
  booktitle    = {{ACL} {(1)}},
  pages        = {1013--1026},
  publisher    = {Association for Computational Linguistics},
  year         = {2019}
}

@inproceedings{DBLP:conf/acl/StabG16,
  author       = {Christian Stab and
                  Iryna Gurevych},
  title        = {Recognizing the Absence of Opposing Arguments in Persuasive Essays},
  booktitle    = {ArgMining@ACL},
  publisher    = {The Association for Computer Linguistics},
  year         = {2016}
}

@inproceedings{DBLP:conf/acl/ReimersSBDSG19,
  author       = {Nils Reimers and
                  Benjamin Schiller and
                  Tilman Beck and
                  Johannes Daxenberger and
                  Christian Stab and
                  Iryna Gurevych},
  title        = {Classification and Clustering of Arguments with Contextualized Word
                  Embeddings},
  booktitle    = {{ACL} {(1)}},
  pages        = {567--578},
  publisher    = {Association for Computational Linguistics},
  year         = {2019}
}

@inproceedings{DBLP:conf/acl/HabernalG16,
  author       = {Ivan Habernal and
                  Iryna Gurevych},
  title        = {Which argument is more convincing? Analyzing and predicting convincingness
                  of Web arguments using bidirectional {LSTM}},
  booktitle    = {{ACL} {(1)}},
  publisher    = {The Association for Computer Linguistics},
  year         = {2016}
}

@inproceedings{DBLP:conf/emnlp/Eckle-KohlerKG15,
  author       = {Judith Eckle{-}Kohler and
                  Roland Kluge and
                  Iryna Gurevych},
  title        = {On the Role of Discourse Markers for Discriminating Claims and Premises
                  in Argumentative Discourse},
  booktitle    = {{EMNLP}},
  pages        = {2236--2242},
  publisher    = {The Association for Computational Linguistics},
  year         = {2015}
}

@inproceedings{DBLP:conf/acl/GleizeSCDMAS19,
  author       = {Martin Gleize and
                  Eyal Shnarch and
                  Leshem Choshen and
                  Lena Dankin and
                  Guy Moshkowich and
                  Ranit Aharonov and
                  Noam Slonim},
  title        = {Are You Convinced? Choosing the More Convincing Evidence with a Siamese
                  Network},
  booktitle    = {{ACL} {(1)}},
  pages        = {967--976},
  publisher    = {Association for Computational Linguistics},
  year         = {2019}
}

@inproceedings{DBLP:conf/coling/LevyBGAS18,
  author       = {Ran Levy and
                  Ben Bogin and
                  Shai Gretz and
                  Ranit Aharonov and
                  Noam Slonim},
  title        = {Towards an argumentative content search engine using weak supervision},
  booktitle    = {{COLING}},
  pages        = {2066--2081},
  publisher    = {Association for Computational Linguistics},
  year         = {2018}
}

@inproceedings{DBLP:conf/eacl/SlonimBSBD17,
  author       = {Roy Bar{-}Haim and
                  Indrajit Bhattacharya and
                  Francesco Dinuzzo and
                  Amrita Saha and
                  Noam Slonim},
  title        = {Stance Classification of Context-Dependent Claims},
  booktitle    = {{EACL} {(1)}},
  pages        = {251--261},
  publisher    = {Association for Computational Linguistics},
  year         = {2017}
}

@inproceedings{DBLP:conf/argmining/FriedmanDHAKS21,
  author       = {Roni Friedman and
                  Lena Dankin and
                  Yufang Hou and
                  Ranit Aharonov and
                  Yoav Katz and
                  Noam Slonim},
  title        = {Overview of the 2021 Key Point Analysis Shared Task},
  booktitle    = {ArgMining@EMNLP},
  pages        = {154--164},
  publisher    = {Association for Computational Linguistics},
  year         = {2021}
}

@inproceedings{DBLP:conf/interspeech/MassSMHSLK18,
  author       = {Yosi Mass and
                  Slava Shechtman and
                  Moran Mordechay and
                  Ron Hoory and
                  Oren Sar Shalom and
                  Guy Lev and
                  David Konopnicki},
  title        = {Word Emphasis Prediction for Expressive Text to Speech},
  booktitle    = {{INTERSPEECH}},
  pages        = {2868--2872},
  publisher    = {{ISCA}},
  year         = {2018}
}

@inproceedings{DBLP:conf/acl/OrbachBTLJAS20,
  author       = {Matan Orbach and
                  Yonatan Bilu and
                  Assaf Toledo and
                  Dan Lahav and
                  Michal Jacovi and
                  Ranit Aharonov and
                  Noam Slonim},
  title        = {Out of the Echo Chamber: Detecting Countering Debate Speeches},
  booktitle    = {{ACL}},
  pages        = {7073--7086},
  publisher    = {Association for Computational Linguistics},
  year         = {2020}
}

@inproceedings{DBLP:conf/emnlp/ShnarchCMAS20,
  author       = {Eyal Shnarch and
                  Leshem Choshen and
                  Guy Moshkowich and
                  Ranit Aharonov and
                  Noam Slonim},
  title        = {Unsupervised Expressive Rules Provide Explainability and Assist Human
                  Experts Grasping New Domains},
  booktitle    = {{EMNLP} (Findings)},
  series       = {Findings of {ACL}},
  volume       = {{EMNLP} 2020},
  pages        = {2678--2697},
  publisher    = {Association for Computational Linguistics},
  year         = {2020}
}

@article{DBLP:journals/coling/LawrenceR19,
  author       = {John Lawrence and
                  Chris Reed},
  title        = {Argument Mining: {A} Survey},
  journal      = {Comput. Linguistics},
  volume       = {45},
  number       = {4},
  pages        = {765--818},
  year         = {2019}
}

@article{patel2024machine,
  title={Machine Learning and Applications in Argumentation Mining.},
  author={Patel, Tavisha},
  journal={International Journal of High School Research},
  volume={6},
  number={1},
  year={2024}
}

@article{DBLP:journals/ipm/LytosLSB19,
  author       = {Anastasios Lytos and
                  Thomas Lagkas and
                  Panagiotis G. Sarigiannidis and
                  Kalina Bontcheva},
  title        = {The evolution of argumentation mining: From models to social media
                  and emerging tools},
  journal      = {Inf. Process. Manag.},
  volume       = {56},
  number       = {6},
  year         = {2019}
}

@inproceedings{DBLP:conf/acl/ChenCLB24,
  author       = {Guizhen Chen and
                  Liying Cheng and
                  Anh Tuan Luu and
                  Lidong Bing},
  title        = {Exploring the Potential of Large Language Models in Computational
                  Argumentation},
  booktitle    = {{ACL} {(1)}},
  pages        = {2309--2330},
  publisher    = {Association for Computational Linguistics},
  year         = {2024}
}

@article{DBLP:journals/corr/abs-2503-00847,
  author       = {Moritz Altemeyer and
                  Steffen Eger and
                  Johannes Daxenberger and
                  Tim Altendorf and
                  Philipp Cimiano and
                  Benjamin Schiller},
  title        = {Argument Summarization and its Evaluation in the Era of Large Language
                  Models},
  journal      = {CoRR},
  volume       = {abs/2503.00847},
  year         = {2025}
}

@article{guida2025llms,
  title={LLMs for Argument Mining: Detection, Extraction, and Relationship Classification of pre-defined Arguments in Online Comments},
  author={Guida, Matteo and Otmakhova, Yulia and Hovy, Eduard and Frermann, Lea},
  journal={arXiv preprint arXiv:2505.22956},
  year={2025}
}

@article{DBLP:journals/ijcini/PeldszusS13,
  author       = {Andreas Peldszus and
                  Manfred Stede},
  title        = {From Argument Diagrams to Argumentation Mining in Texts: {A} Survey},
  journal      = {Int. J. Cogn. Informatics Nat. Intell.},
  volume       = {7},
  number       = {1},
  pages        = {1--31},
  year         = {2013}
}

@inproceedings{DBLP:conf/ijcai/CabrioV18,
  author       = {Elena Cabrio and
                  Serena Villata},
  title        = {Five Years of Argument Mining: a Data-driven Analysis},
  booktitle    = {{IJCAI}},
  pages        = {5427--5433},
  publisher    = {ijcai.org},
  year         = {2018}
}

@inproceedings{DBLP:conf/naacl/HuaNBW19,
  author       = {Xinyu Hua and
                  Mitko Nikolov and
                  Nikhil Badugu and
                  Lu Wang},
  title        = {Argument Mining for Understanding Peer Reviews},
  booktitle    = {{NAACL-HLT} {(1)}},
  pages        = {2131--2137},
  publisher    = {Association for Computational Linguistics},
  year         = {2019}
}

@inproceedings{DBLP:conf/clic-it/GrundlerGSFGPLS24,
  author       = {Giulia Grundler and
                  Andrea Galassi and
                  Piera Santin and
                  Alessia Fidelangeli and
                  Federico Galli and
                  Elena Palmieri and
                  Francesca Lagioia and
                  Giovanni Sartor and
                  Paolo Torroni},
  title        = {{AMELIA} - Argument Mining Evaluation on Legal documents in ItAlian:
                  {A} {CALAMITA} Challenge},
  booktitle    = {CLiC-it},
  series       = {{CEUR} Workshop Proceedings},
  volume       = {3878},
  publisher    = {CEUR-WS.org},
  year         = {2024}
}

@inproceedings{DBLP:conf/comma/RuoschLSB24,
  author       = {Florian Ruosch and
                  John Lawrence and
                  Cristina Sarasua and
                  Abraham Bernstein},
  title        = {A Unified Benchmark for Argument Mining},
  booktitle    = {{COMMA}},
  series       = {Frontiers in Artificial Intelligence and Applications},
  volume       = {388},
  pages        = {363--364},
  publisher    = {{IOS} Press},
  year         = {2024}
}

@inproceedings{DBLP:conf/emnlp/SviridovaYECVA24,
  author       = {Ekaterina Sviridova and
                  Anar Yeginbergen and
                  Ainara Estarrona and
                  Elena Cabrio and
                  Serena Villata and
                  Rodrigo Agerri},
  title        = {CasiMedicos-Arg: {A} Medical Question Answering Dataset Annotated
                  with Explanatory Argumentative Structures},
  booktitle    = {{EMNLP}},
  pages        = {18463--18475},
  publisher    = {Association for Computational Linguistics},
  year         = {2024}
}

@phdthesis{DBLP:phd/dnb/Alhamzeh23,
  author       = {Alaa Alhamzeh},
  title        = {Language Reasoning by means of Argument Mining and Argument Quality},
  school       = {University of Passau, Germany},
  year         = {2023}
}

@article{DBLP:journals/taslp/GalassiLT23,
  author       = {Andrea Galassi and
                  Marco Lippi and
                  Paolo Torroni},
  title        = {Multi-Task Attentive Residual Networks for Argument Mining},
  journal      = {{IEEE} {ACM} Trans. Audio Speech Lang. Process.},
  volume       = {31},
  pages        = {1877--1892},
  year         = {2023}
}

@inproceedings{DBLP:conf/argmining/LiuEZL23,
  author       = {Zhexiong Liu and
                  Mohamed Elaraby and
                  Yang Zhong and
                  Diane J. Litman},
  title        = {Overview of ImageArg-2023: The First Shared Task in Multimodal Argument
                  Mining},
  booktitle    = {ArgMining@EMNLP},
  pages        = {120--132},
  publisher    = {Association for Computational Linguistics},
  year         = {2023}
}

@inproceedings{DBLP:conf/emnlp/Ruiz-DolzS23,
  author       = {Ramon Ruiz{-}Dolz and
                  Javier Sanchez},
  title        = {VivesDebate-Speech: {A} Corpus of Spoken Argumentation to Leverage
                  Audio Features for Argument Mining},
  booktitle    = {{EMNLP}},
  pages        = {2071--2077},
  publisher    = {Association for Computational Linguistics},
  year         = {2023}
}

@inproceedings{DBLP:conf/icail/ZhangNL23,
  author       = {Gechuan Zhang and
                  Paul Nulty and
                  David Lillis},
  title        = {Argument Mining with Graph Representation Learning},
  booktitle    = {{ICAIL}},
  pages        = {371--380},
  publisher    = {{ACM}},
  year         = {2023}
}

@article{DBLP:journals/corr/abs-2206-09249,
  author       = {Evgeny V. Kotelnikov and
                  Natalia V. Loukachevitch and
                  Irina Nikishina and
                  Alexander Panchenko},
  title        = {RuArg-2022: Argument Mining Evaluation},
  journal      = {CoRR},
  volume       = {abs/2206.09249},
  year         = {2022}
}

@article{DBLP:journals/it/SchaeferS21,
  author       = {Robin Schaefer and
                  Manfred Stede},
  title        = {Argument Mining on Twitter: {A} survey},
  journal      = {it Inf. Technol.},
  volume       = {63},
  number       = {1},
  pages        = {45--58},
  year         = {2021}
}

@inproceedings{DBLP:conf/aaai/0001FBBQZDSM021,
  author       = {Michael Fromm and
                  Evgeniy Faerman and
                  Max Berrendorf and
                  Siddharth Bhargava and
                  Ruoxia Qi and
                  Yao Zhang and
                  Lukas Dennert and
                  Sophia Selle and
                  Yang Mao and
                  Thomas Seidl},
  title        = {Argument Mining Driven Analysis of Peer-Reviews},
  booktitle    = {{AAAI}},
  pages        = {4758--4766},
  publisher    = {{AAAI} Press},
  year         = {2021}
}

@inproceedings{DBLP:conf/aaai/Ein-DorSDHSGAGC20,
  author       = {Liat Ein{-}Dor and
                  Eyal Shnarch and
                  Lena Dankin and
                  Alon Halfon and
                  Benjamin Sznajder and
                  Ariel Gera and
                  Carlos Alzate and
                  Martin Gleize and
                  Leshem Choshen and
                  Yufang Hou and
                  Yonatan Bilu and
                  Ranit Aharonov and
                  Noam Slonim},
  title        = {Corpus Wide Argument Mining - {A} Working Solution},
  booktitle    = {{AAAI}},
  pages        = {7683--7691},
  publisher    = {{AAAI} Press},
  year         = {2020}
}

@inproceedings{DBLP:conf/ecai/0002CV20,
  author       = {Tobias Mayer and
                  Elena Cabrio and
                  Serena Villata},
  title        = {Transformer-Based Argument Mining for Healthcare Applications},
  booktitle    = {{ECAI}},
  series       = {Frontiers in Artificial Intelligence and Applications},
  volume       = {325},
  pages        = {2108--2115},
  publisher    = {{IOS} Press},
  year         = {2020}
}

@article{long2024llms,
  title={On llms-driven synthetic data generation, curation, and evaluation: A survey},
  author={Long, Lin and Wang, Rui and Xiao, Ruixuan and Zhao, Junbo and Ding, Xiao and Chen, Gang and Wang, Haobo},
  journal={arXiv preprint arXiv:2406.15126},
  year={2024}
}

@inproceedings{DBLP:conf/lrec/PersingN20,
  author       = {Isaac Persing and
                  Vincent Ng},
  title        = {Unsupervised Argumentation Mining in Student Essays},
  booktitle    = {{LREC}},
  pages        = {6795--6803},
  publisher    = {European Language Resources Association},
  year         = {2020}
}

@article{DBLP:journals/corr/abs-2304-07666,
  author       = {Yikang Liu and
                  Ziyin Zhang and
                  Wanyang Zhang and
                  Shisen Yue and
                  Xiaojing Zhao and
                  Xinyuan Cheng and
                  Yiwen Zhang and
                  Hai Hu},
  title        = {ArguGPT: evaluating, understanding and identifying argumentative essays
                  generated by {GPT} models},
  journal      = {CoRR},
  volume       = {abs/2304.07666},
  year         = {2023}
}

@inproceedings{DBLP:conf/emnlp/ChengBYLS20,
  author       = {Liying Cheng and
                  Lidong Bing and
                  Qian Yu and
                  Wei Lu and
                  Luo Si},
  title        = {{APE:} Argument Pair Extraction from Peer Review and Rebuttal via
                  Multi-task Learning},
  booktitle    = {{EMNLP} {(1)}},
  pages        = {7000--7011},
  publisher    = {Association for Computational Linguistics},
  year         = {2020}
}

@inproceedings{DBLP:conf/naacl/ChenK0CR19,
  author       = {Sihao Chen and
                  Daniel Khashabi and
                  Wenpeng Yin and
                  Chris Callison{-}Burch and
                  Dan Roth},
  title        = {Seeing Things from a Different Angle: Discovering Diverse Perspectives
                  about Claims},
  booktitle    = {{NAACL-HLT} {(1)}},
  pages        = {542--557},
  publisher    = {Association for Computational Linguistics},
  year         = {2019}
}

@inproceedings{DBLP:conf/argmining/HuberTRDB19,
  author       = {Laurine Huber and
                  Yannick Toussaint and
                  Charlotte Roze and
                  Mathilde Dargnat and
                  Chlo{\'{e}} Braud},
  title        = {Aligning Discourse and Argumentation Structures using Subtrees and
                  Redescription Mining},
  booktitle    = {ArgMining@ACL},
  pages        = {35--40},
  publisher    = {Association for Computational Linguistics},
  year         = {2019}
}

@inproceedings{DBLP:conf/acl/Bar-HaimEFKLS20,
  author       = {Roy Bar{-}Haim and
                  Lilach Eden and
                  Roni Friedman and
                  Yoav Kantor and
                  Dan Lahav and
                  Noam Slonim},
  title        = {From Arguments to Key Points: Towards Automatic Argument Summarization},
  booktitle    = {{ACL}},
  pages        = {4029--4039},
  publisher    = {Association for Computational Linguistics},
  year         = {2020}
}

@inproceedings{DBLP:conf/acl/Tang0DC24,
  author       = {An Quang Tang and
                  Xiuzhen Zhang and
                  Minh Ngoc Dinh and
                  Erik Cambria},
  title        = {Prompted Aspect Key Point Analysis for Quantitative Review Summarization},
  booktitle    = {{ACL} {(1)}},
  pages        = {10691--10708},
  publisher    = {Association for Computational Linguistics},
  year         = {2024}
}

@inproceedings{DBLP:conf/eacl/TangZD24,
  author       = {An Quang Tang and
                  Xiuzhen Zhang and
                  Minh Ngoc Dinh},
  title        = {Aspect-based Key Point Analysis for Quantitative Summarization of
                  Reviews},
  booktitle    = {{EACL} (Findings)},
  pages        = {1419--1433},
  publisher    = {Association for Computational Linguistics},
  year         = {2024}
}

@inproceedings{DBLP:conf/naacl/LiJHXCH24,
  author       = {Xiao Li and
                  Yong Jiang and
                  Shen Huang and
                  Pengjun Xie and
                  Gong Cheng and
                  Fei Huang},
  title        = {Exploring Key Point Analysis with Pairwise Generation and Graph Partitioning},
  booktitle    = {{NAACL-HLT}},
  pages        = {5657--5667},
  publisher    = {Association for Computational Linguistics},
  year         = {2024}
}

@article{DBLP:journals/corr/abs-2404-18371,
  author       = {Tomoki Fukuma and
                  Koki Noda and
                  Toshihide Ubukata and
                  Kousuke Hosoi and
                  Yoshiharu Ichikawa and
                  Kyosuke Kambe and
                  Yu Masubuchi and
                  Fujio Toriumi},
  title        = {{QANA:} LLM-based Question Generation and Network Analysis for Zero-shot
                  Key Point Analysis and Beyond},
  journal      = {CoRR},
  volume       = {abs/2404.18371},
  year         = {2024}
}

@inproceedings{DBLP:conf/acl/LiSBN23,
  author       = {Hao Li and
                  Viktor Schlegel and
                  Riza Batista{-}Navarro and
                  Goran Nenadic},
  title        = {Do You Hear The People Sing? Key Point Analysis via Iterative Clustering
                  and Abstractive Summarisation},
  booktitle    = {{ACL} {(1)}},
  pages        = {14064--14080},
  publisher    = {Association for Computational Linguistics},
  year         = {2023}
}

@inproceedings{DBLP:conf/emnlp/EdenKOKSB23,
  author       = {Lilach Eden and
                  Yoav Kantor and
                  Matan Orbach and
                  Yoav Katz and
                  Noam Slonim and
                  Roy Bar{-}Haim},
  title        = {Welcome to the Real World: Efficient, Incremental and Scalable Key
                  Point Analysis},
  booktitle    = {{EMNLP} (Industry Track)},
  pages        = {483--491},
  publisher    = {Association for Computational Linguistics},
  year         = {2023}
}

@inproceedings{DBLP:conf/acl/Bar-HaimEKFS20,
  author       = {Roy Bar{-}Haim and
                  Lilach Eden and
                  Yoav Kantor and
                  Roni Friedman and
                  Noam Slonim},
  title        = {Every Bite Is an Experience: Key Point Analysis of Business Reviews},
  booktitle    = {{ACL/IJCNLP} {(1)}},
  pages        = {3376--3386},
  publisher    = {Association for Computational Linguistics},
  year         = {2021}
}

@inproceedings{DBLP:conf/argmining/AlshomaryGSHSCP21,
  author       = {Milad Alshomary and
                  Timon Gurcke and
                  Shahbaz Syed and
                  Philipp Heinisch and
                  Maximilian Splieth{\"{o}}ver and
                  Philipp Cimiano and
                  Martin Potthast and
                  Henning Wachsmuth},
  title        = {Key Point Analysis via Contrastive Learning and Extractive Argument
                  Summarization},
  booktitle    = {ArgMining@EMNLP},
  pages        = {184--189},
  publisher    = {Association for Computational Linguistics},
  year         = {2021}
}

@inproceedings{DBLP:conf/argmining/PhanNND21,
  author       = {Viet Hoang Phan and
                  Tien Long Nguyen and
                  Duc Long Nguyen and
                  Ngoc Khanh Doan},
  title        = {Matching The Statements: {A} Simple and Accurate Model for Key Point
                  Analysis},
  booktitle    = {ArgMining@EMNLP},
  pages        = {165--174},
  publisher    = {Association for Computational Linguistics},
  year         = {2021}
}

@inproceedings{DBLP:conf/argmining/ReimerLHA21,
  author       = {Jan Heinrich Reimer and
                  Thi Kim Hanh Luu and
                  Max Henze and
                  Yamen Ajjour},
  title        = {Modern Talking in Key Point Analysis: Key Point Matching using Pretrained
                  Encoders},
  booktitle    = {ArgMining@EMNLP},
  pages        = {175--183},
  publisher    = {Association for Computational Linguistics},
  year         = {2021}
}

@article{DBLP:journals/ijcse/GaoGG19,
  author       = {Yongchang Gao and
                  Haowen Guan and
                  Bin Gong},
  title        = {{CODM:} an outlier detection method for medical insurance claims fraud},
  journal      = {Int. J. Comput. Sci. Eng.},
  volume       = {20},
  number       = {3},
  pages        = {404--411},
  year         = {2019}
}

@inproceedings{DBLP:conf/icacds/MohanP19,
  author       = {Thanusree Mohan and
                  K. Praveen},
  title        = {Fraud Detection in Medical Insurance Claim with Privacy Preserving
                  Data Publishing in {TLS-N} Using Blockchain},
  booktitle    = {{ICACDS} {(1)}},
  series       = {Communications in Computer and Information Science},
  volume       = {1045},
  pages        = {211--220},
  publisher    = {Springer},
  year         = {2019}
}

@inproceedings{DBLP:conf/acl/DayanikP20,
  author       = {Erenay Dayanik and
                  Sebastian Pad{\'{o}}},
  title        = {Masking Actor Information Leads to Fairer Political Claims Detection},
  booktitle    = {{ACL}},
  pages        = {4385--4391},
  publisher    = {Association for Computational Linguistics},
  year         = {2020}
}

@inproceedings{DBLP:conf/cikm/ChenHC20,
  author       = {Chung{-}Chi Chen and
                  Hen{-}Hsen Huang and
                  Hsin{-}Hsi Chen},
  title        = {NumClaim: Investor's Fine-grained Claim Detection},
  booktitle    = {{CIKM}},
  pages        = {1973--1976},
  publisher    = {{ACM}},
  year         = {2020}
}

@inproceedings{DBLP:conf/edm/WanCAM20,
  author       = {Qian Wan and
                  Scott A. Crossley and
                  Laura K. Allen and
                  Danielle S. McNamara},
  title        = {Claim Detection and Relationship with Writing Quality},
  booktitle    = {{EDM}},
  publisher    = {International Educational Data Mining Society},
  year         = {2020}
}

@inproceedings{DBLP:conf/emnlp/WrightA20,
  author       = {Dustin Wright and
                  Isabelle Augenstein},
  title        = {Claim Check-Worthiness Detection as Positive Unlabelled Learning},
  booktitle    = {{EMNLP} (Findings)},
  series       = {Findings of {ACL}},
  volume       = {{EMNLP} 2020},
  pages        = {476--488},
  publisher    = {Association for Computational Linguistics},
  year         = {2020}
}

@article{DBLP:journals/osnm/BerendtBHJPA21,
  author       = {Bettina Berendt and
                  Peter Burger and
                  Rafael Hautekiet and
                  Jan Jagers and
                  Alexander Pleijter and
                  Peter Van Aelst},
  title        = {FactRank: Developing automated claim detection for Dutch-language
                  fact-checkers},
  journal      = {Online Soc. Networks Media},
  volume       = {22},
  pages        = {100113},
  year         = {2021}
}

@inproceedings{DBLP:conf/bionlp/WuhrlK21,
  author       = {Amelie W{\"{u}}hrl and
                  Roman Klinger},
  title        = {Claim Detection in Biomedical Twitter Posts},
  booktitle    = {BioNLP@NAACL-HLT},
  pages        = {131--142},
  publisher    = {Association for Computational Linguistics},
  year         = {2021}
}

@inproceedings{DBLP:conf/emnlp/LinMCYCC21,
  author       = {Hongzhan Lin and
                  Jing Ma and
                  Mingfei Cheng and
                  Zhiwei Yang and
                  Liangliang Chen and
                  Guang Chen},
  title        = {Rumor Detection on Twitter with Claim-Guided Hierarchical Graph Attention
                  Networks},
  booktitle    = {{EMNLP} {(1)}},
  pages        = {10035--10047},
  publisher    = {Association for Computational Linguistics},
  year         = {2021}
}

@inproceedings{DBLP:conf/fire/WoloszynKS21,
  author       = {Vinicius Woloszyn and
                  Joseph Kobti and
                  Vera Schmitt},
  title        = {Towards Automatic Green Claim Detection},
  booktitle    = {{FIRE}},
  pages        = {28--34},
  publisher    = {{ACM}},
  year         = {2021}
}

@inproceedings{DBLP:conf/infrkm/YaziVRTL21,
  author       = {Fatin Syafiqah Yazi and
                  Wan{-}Tze Vong and
                  Valliappan Raman and
                  Patrick Hang Hui Then and
                  Mukulraj J. Lunia},
  title        = {Towards Automated Detection of Contradictory Research Claims in Medical
                  Literature Using Deep Learning Approach},
  booktitle    = {{CAMP}},
  pages        = {116--121},
  publisher    = {{IEEE}},
  year         = {2021}
}

@inproceedings{DBLP:conf/www/BeltranML21,
  author       = {Javier Beltr{\'{a}}n and
                  Rub{\'{e}}n M{\'{\i}}guez and
                  Irene Larraz},
  title        = {ClaimHunter: An Unattended Tool for Automated Claim Detection on Twitter},
  booktitle    = {KnOD@WWW},
  series       = {{CEUR} Workshop Proceedings},
  volume       = {2877},
  publisher    = {CEUR-WS.org},
  year         = {2021}
}

@inproceedings{DBLP:conf/naacl/CheemaHSMOE22,
  author       = {Gullal Singh Cheema and
                  Sherzod Hakimov and
                  Abdul Sittar and
                  Eric M{\"{u}}ller{-}Budack and
                  Christian Otto and
                  Ralph Ewerth},
  title        = {MM-Claims: {A} Dataset for Multimodal Claim Detection in Social Media},
  booktitle    = {{NAACL-HLT} (Findings)},
  pages        = {962--979},
  publisher    = {Association for Computational Linguistics},
  year         = {2022}
}

@inproceedings{DBLP:conf/acl/StammbachWBKL23,
  author       = {Dominik Stammbach and
                  Nicolas Webersinke and
                  Julia Anna Bingler and
                  Mathias Kraus and
                  Markus Leippold},
  title        = {Environmental Claim Detection},
  booktitle    = {{ACL} {(2)}},
  pages        = {1051--1066},
  publisher    = {Association for Computational Linguistics},
  year         = {2023}
}

@inproceedings{DBLP:conf/ecir/SchlichtFR23,
  author       = {Ipek Baris Schlicht and
                  Lucie Flek and
                  Paolo Rosso},
  title        = {Multilingual Detection of Check-Worthy Claims Using World Languages
                  and Adapter Fusion},
  booktitle    = {{ECIR} {(1)}},
  series       = {Lecture Notes in Computer Science},
  volume       = {13980},
  pages        = {118--133},
  publisher    = {Springer},
  year         = {2023}
}

@inproceedings{DBLP:conf/fire/SundriyalA023,
  author       = {Megha Sundriyal and
                  Md. Shad Akhtar and
                  Tanmoy Chakraborty},
  title        = {Overview of the {CLAIMSCAN-2023:} Uncovering Truth in Social Media
                  through Claim Detection and Identification of Claim Spans},
  booktitle    = {{FIRE}},
  pages        = {7--9},
  publisher    = {{ACM}},
  year         = {2023}
}

@phdthesis{DBLP:phd/hal/Hafid24,
  author       = {Salim Hafid},
  title        = {Detection, linking and interpretation of science-related claims and
                  their contexts from online discourse. (D{\'{e}}tection, liage, et
                  interpr{\'{e}}tation d'{\'{e}}nonc{\'{e}}s scientifiques et leurs
                  contextes {\`{a}} partir de discours en ligne)},
  school       = {University of Montpellier, France},
  year         = {2024}
}

@inproceedings{DBLP:conf/visigrapp/Rayar24a,
  author       = {Fr{\'{e}}d{\'{e}}ric Rayar},
  title        = {Fact-Checked Claim Detection in Videos Using a Multimodal Approach},
  booktitle    = {{VISIGRAPP} {(4):} {VISAPP}},
  pages        = {614--620},
  publisher    = {{SCITEPRESS}},
  year         = {2024}
}

@article{DBLP:journals/corr/abs-2402-11728,
  author       = {Agam Shah and
                  Arnav Hiray and
                  Pratvi Shah and
                  Arkaprabha Banerjee and
                  Anushka Singh and
                  Dheeraj Eidnani and
                  Bhaskar Chaudhury and
                  Sudheer Chava},
  title        = {Numerical Claim Detection in Finance: {A} New Financial Dataset, Weak-Supervision
                  Model, and Market Analysis},
  journal      = {CoRR},
  volume       = {abs/2402.11728},
  year         = {2024}
}

@article{DBLP:journals/corr/abs-2407-18367,
  author       = {Nazanin Jafari and
                  James Allan},
  title        = {Robust Claim Verification Through Fact Detection},
  journal      = {CoRR},
  volume       = {abs/2407.18367},
  year         = {2024}
}

@article{DBLP:journals/artmed/PreezBBB25,
  author       = {Anli du Preez and
                  Sanmitra Bhattacharya and
                  Peter A. Beling and
                  Edward Bowen},
  title        = {Fraud detection in healthcare claims using machine learning: {A} systematic
                  review},
  journal      = {Artif. Intell. Medicine},
  volume       = {160},
  pages        = {103061},
  year         = {2025}
}

@article{DBLP:journals/kbs/TubishatTAHH25,
  author       = {Mohammad Tubishat and
                  Dina Tbaishat and
                  Ala' M. Al{-}Zoubi and
                  Abed{-}Elalim Hraiz and
                  Maria Habib},
  title        = {Leveraging evolutionary algorithms with a dynamic weighted search
                  space approach for fraud detection in healthcare insurance claims},
  journal      = {Knowl. Based Syst.},
  volume       = {317},
  pages        = {113436},
  year         = {2025}
}

@article{DBLP:journals/corr/abs-2503-02737,
  author       = {Ivan Vykopal and
                  Mat{\'{u}}s Pikuliak and
                  Simon Ostermann and
                  Tatiana Anikina and
                  Michal Gregor and
                  Mari{\'{a}}n Simko},
  title        = {Large Language Models for Multilingual Previously Fact-Checked Claim
                  Detection},
  journal      = {CoRR},
  volume       = {abs/2503.02737},
  year         = {2025}
}

@article{DBLP:journals/corr/abs-2503-15220,
  author       = {Rrubaa Panchendrarajan and
                  Arkaitz Zubiaga},
  title        = {Entity-aware Cross-lingual Claim Detection for Automated Fact-checking},
  journal      = {CoRR},
  volume       = {abs/2503.15220},
  year         = {2025}
}

@article{DBLP:journals/corr/abs-2504-12882,
  author       = {Patrick Giedemann and
                  Pius von D{\"{a}}niken and
                  Jan Deriu and
                  {\'{A}}lvaro Rodrigo and
                  Anselmo Pe{\~{n}}as and
                  Mark Cieliebak},
  title        = {ViClaim: {A} Multilingual Multilabel Dataset for Automatic Claim Detection
                  in Videos},
  journal      = {CoRR},
  volume       = {abs/2504.12882},
  year         = {2025}
}

@article{DBLP:journals/eaai/HuangMRJS25,
  author       = {Xuejian Huang and
                  Tinghuai Ma and
                  Huan Rong and
                  Li Jia and
                  Yuming Su},
  title        = {Dual evidence enhancement and text-image similarity awareness for
                  multimodal rumor detection},
  journal      = {Eng. Appl. Artif. Intell.},
  volume       = {153},
  pages        = {110845},
  year         = {2025}
}

@article{DBLP:journals/fi/FerdushKKGD25,
  author       = {Jannatul Ferdush and
                  Joarder Kamruzzaman and
                  Gour C. Karmakar and
                  Iqbal Gondal and
                  Rajkumar Das},
  title        = {Cross-Domain Fake News Detection Through Fusion of Evidence from Multiple
                  Social Media Platforms},
  journal      = {Future Internet},
  volume       = {17},
  number       = {2},
  pages        = {61},
  year         = {2025}
}

@article{DBLP:journals/tkde/LiuWWW24,
  author       = {Qiang Liu and
                  Junfei Wu and
                  Shu Wu and
                  Liang Wang},
  title        = {Out-of-Distribution Evidence-Aware Fake News Detection via Dual Adversarial
                  Debiasing},
  journal      = {{IEEE} Trans. Knowl. Data Eng.},
  volume       = {36},
  number       = {11},
  pages        = {6801--6813},
  year         = {2024}
}

@inproceedings{DBLP:conf/emnlp/JiayangCZQZLS0L24,
  author       = {Cheng Jiayang and
                  Chunkit Chan and
                  Qianqian Zhuang and
                  Lin Qiu and
                  Tianhang Zhang and
                  Tengxiao Liu and
                  Yangqiu Song and
                  Yue Zhang and
                  Pengfei Liu and
                  Zheng Zhang},
  title        = {{ECON:} On the Detection and Resolution of Evidence Conflicts},
  booktitle    = {{EMNLP}},
  pages        = {7816--7844},
  publisher    = {Association for Computational Linguistics},
  year         = {2024}
}

@inproceedings{DBLP:conf/icassp/WuC24,
  author       = {Kaixuan Wu and
                  Donglin Cao},
  title        = {Evidence-Aware Multimodal Chinese Social Media Rumor Detection},
  booktitle    = {{ICASSP}},
  pages        = {8376--8380},
  publisher    = {{IEEE}},
  year         = {2024}
}

@inproceedings{DBLP:conf/ijcai/WuWZ24,
  author       = {Lianwei Wu and
                  Linyong Wang and
                  Yongqiang Zhao},
  title        = {Unified Evidence Enhancement Inference Framework for Fake News Detection},
  booktitle    = {{IJCAI}},
  pages        = {6541--6549},
  publisher    = {ijcai.org},
  year         = {2024}
}

@article{DBLP:journals/corr/abs-2407-01213,
  author       = {Qingxing Dong and
                  Mengyi Zhang and
                  Shiyuan Wu and
                  Xiaozhen Wu},
  title        = {{EMIF:} Evidence-aware Multi-source Information Fusion Network for
                  Explainable Fake News Detection},
  journal      = {CoRR},
  volume       = {abs/2407.01213},
  year         = {2024}
}

@article{DBLP:journals/jimaging/DementievaKP23,
  author       = {Daryna Dementieva and
                  Mikhail Kuimov and
                  Alexander Panchenko},
  title        = {Multiverse: Multilingual Evidence for Fake News Detection},
  journal      = {J. Imaging},
  volume       = {9},
  number       = {4},
  pages        = {77},
  year         = {2023}
}

@inproceedings{DBLP:conf/cikm/GuoZT023,
  author       = {Hao Guo and
                  Weixin Zeng and
                  Jiuyang Tang and
                  Xiang Zhao},
  title        = {Interpretable Fake News Detection with Graph Evidence},
  booktitle    = {{CIKM}},
  pages        = {659--668},
  publisher    = {{ACM}},
  year         = {2023}
}

@inproceedings{DBLP:conf/kdd/LiaoPHZLSX23,
  author       = {Hao Liao and
                  Jiahao Peng and
                  Zhanyi Huang and
                  Wei Zhang and
                  Guanghua Li and
                  Kai Shu and
                  Xing Xie},
  title        = {{MUSER:} {A} MUlti-Step Evidence Retrieval Enhancement Framework for
                  Fake News Detection},
  booktitle    = {{KDD}},
  pages        = {4461--4472},
  publisher    = {{ACM}},
  year         = {2023}
}

@article{DBLP:journals/corr/abs-2306-13450,
  author       = {Hao Liao and
                  Jiaohao Peng and
                  Zhanyi Huang and
                  Wei Zhang and
                  Guanghua Li and
                  Kai Shu and
                  Xing Xie},
  title        = {{MUSER:} {A} MUlti-Step Evidence Retrieval Enhancement Framework for
                  Fake News Detection},
  journal      = {CoRR},
  volume       = {abs/2306.13450},
  year         = {2023}
}

@article{DBLP:journals/access/HammouchiG22,
  author       = {Hicham Hammouchi and
                  Mounir Ghogho},
  title        = {Evidence-Aware Multilingual Fake News Detection},
  journal      = {{IEEE} Access},
  volume       = {10},
  pages        = {116808--116818},
  year         = {2022}
}

@inproceedings{DBLP:conf/iccci/KharratHR22,
  author       = {Ala Eddine Kharrat and
                  Lobna Hlaoua and
                  Lotfi Ben Romdhane},
  title        = {Contradiction Detection Approach Based on Semantic Relations and Evidence
                  of Uncertainty},
  booktitle    = {{ICCCI}},
  series       = {Lecture Notes in Computer Science},
  volume       = {13501},
  pages        = {232--245},
  publisher    = {Springer},
  year         = {2022}
}

@inproceedings{DBLP:conf/www/XuWLWW22,
  author       = {Weizhi Xu and
                  Junfei Wu and
                  Qiang Liu and
                  Shu Wu and
                  Liang Wang},
  title        = {Evidence-aware Fake News Detection with Graph Neural Networks},
  booktitle    = {{WWW}},
  pages        = {2501--2510},
  publisher    = {{ACM}},
  year         = {2022}
}

@article{DBLP:journals/corr/abs-2201-06885,
  author       = {Weizhi Xu and
                  Junfei Wu and
                  Qiang Liu and
                  Shu Wu and
                  Liang Wang},
  title        = {Mining Fine-grained Semantics via Graph Neural Networks for Evidence-based
                  Fake News Detection},
  journal      = {CoRR},
  volume       = {abs/2201.06885},
  year         = {2022}
}

@inproceedings{DBLP:conf/acl/DementievaP21,
  author       = {Daryna Dementieva and
                  Alexander Panchenko},
  title        = {Cross-lingual Evidence Improves Monolingual Fake News Detection},
  booktitle    = {{ACL} (student)},
  pages        = {310--320},
  publisher    = {Association for Computational Linguistics},
  year         = {2021}
}

@inproceedings{DBLP:conf/argmining/ElarabyL21,
  author       = {Mohamed Elaraby and
                  Diane J. Litman},
  title        = {Self-trained Pretrained Language Models for Evidence Detection},
  booktitle    = {ArgMining@EMNLP},
  pages        = {142--147},
  publisher    = {Association for Computational Linguistics},
  year         = {2021}
}

@inproceedings{DBLP:conf/dsaa/DementievaP20,
  author       = {Daryna Dementieva and
                  Alexander Panchenko},
  title        = {Fake News Detection using Multilingual Evidence},
  booktitle    = {{DSAA}},
  pages        = {775--776},
  publisher    = {{IEEE}},
  year         = {2020}
}

@article{DBLP:journals/ijon/ShanL25,
  author       = {Huawei Shan and
                  Dongyuan Lu},
  title        = {Debate divides: Argument relation-based contrastive opinion summarization
                  via multi-task learning for online discussions},
  journal      = {Neurocomputing},
  volume       = {638},
  pages        = {130124},
  year         = {2025}
}

@inproceedings{DBLP:conf/eacl/MeerVJM24,
  author       = {Michiel van der Meer and
                  Piek Vossen and
                  Catholijn M. Jonker and
                  Pradeep K. Murukannaiah},
  title        = {An Empirical Analysis of Diversity in Argument Summarization},
  booktitle    = {{EACL} {(1)}},
  pages        = {2028--2045},
  publisher    = {Association for Computational Linguistics},
  year         = {2024}
}

@inproceedings{DBLP:conf/naacl/Khosravani0T24,
  author       = {Mohammad Khosravani and
                  Chenyang Huang and
                  Amine Trabelsi},
  title        = {Enhancing Argument Summarization: Prioritizing Exhaustiveness in Key
                  Point Generation and Introducing an Automatic Coverage Evaluation
                  Metric},
  booktitle    = {{NAACL-HLT}},
  pages        = {8212--8224},
  publisher    = {Association for Computational Linguistics},
  year         = {2024}
}

@inproceedings{DBLP:conf/nips/RoushSBZMZBVS24,
  author       = {Allen Roush and
                  Yusuf Shabazz and
                  Arvind Balaji and
                  Peter Zhang and
                  Stefano Mezza and
                  Markus Zhang and
                  Sanjay Basu and
                  Sriram Vishwanath and
                  Ravid Shwartz{-}Ziv},
  title        = {OpenDebateEvidence: {A} Massive-Scale Argument Mining and Summarization
                  Dataset},
  booktitle    = {NeurIPS},
  year         = {2024}
}

@inproceedings{DBLP:conf/acl/ElarabyZL23,
  author       = {Mohamed Elaraby and
                  Yang Zhong and
                  Diane J. Litman},
  title        = {Towards Argument-Aware Abstractive Summarization of Long Legal Opinions
                  with Summary Reranking},
  booktitle    = {{ACL} (Findings)},
  pages        = {7601--7612},
  publisher    = {Association for Computational Linguistics},
  year         = {2023}
}

@inproceedings{DBLP:conf/emnlp/ZhaoWP23,
  author       = {Xiutian Zhao and
                  Ke Wang and
                  Wei Peng},
  title        = {{ORCHID:} {A} Chinese Debate Corpus for Target-Independent Stance
                  Detection and Argumentative Dialogue Summarization},
  booktitle    = {{EMNLP}},
  pages        = {9358--9375},
  publisher    = {Association for Computational Linguistics},
  year         = {2023}
}

@inproceedings{DBLP:conf/icail/XuA23,
  author       = {Huihui Xu and
                  Kevin D. Ashley},
  title        = {Argumentative Segmentation Enhancement for Legal Summarization},
  booktitle    = {ASAIL@ICAIL},
  series       = {{CEUR} Workshop Proceedings},
  volume       = {3441},
  pages        = {141--150},
  publisher    = {CEUR-WS.org},
  year         = {2023}
}

@inproceedings{DBLP:conf/sigdial/SyedZHWP23,
  author       = {Shahbaz Syed and
                  Timon Ziegenbein and
                  Philipp Heinisch and
                  Henning Wachsmuth and
                  Martin Potthast},
  title        = {Frame-oriented Summarization of Argumentative Discussions},
  booktitle    = {{SIGDIAL}},
  pages        = {114--129},
  publisher    = {Association for Computational Linguistics},
  year         = {2023}
}

@inproceedings{DBLP:conf/coling/ElarabyL22,
  author       = {Mohamed Elaraby and
                  Diane J. Litman},
  title        = {ArgLegalSumm: Improving Abstractive Summarization of Legal Documents
                  with Argument Mining},
  booktitle    = {{COLING}},
  pages        = {6187--6194},
  publisher    = {International Committee on Computational Linguistics},
  year         = {2022}
}

@inproceedings{DBLP:conf/jurix/SteffesR22,
  author       = {Bianca Steffes and
                  Piotr Ryszard Rataj},
  title        = {Legal Text Summarization Using Argumentative Structures},
  booktitle    = {{JURIX}},
  series       = {Frontiers in Artificial Intelligence and Applications},
  volume       = {362},
  pages        = {243--248},
  publisher    = {{IOS} Press},
  year         = {2022}
}

@inproceedings{DBLP:conf/acl/FabbriRRWLMR20,
  author       = {Alexander R. Fabbri and
                  Faiaz Rahman and
                  Imad Rizvi and
                  Borui Wang and
                  Haoran Li and
                  Yashar Mehdad and
                  Dragomir R. Radev},
  title        = {ConvoSumm: Conversation Summarization Benchmark and Improved Abstractive
                  Summarization with Argument Mining},
  booktitle    = {{ACL/IJCNLP} {(1)}},
  pages        = {6866--6880},
  publisher    = {Association for Computational Linguistics},
  year         = {2021}
}

@article{DBLP:journals/corr/abs-2503-00024,
  author       = {Yanran Chen and
                  Steffen Eger},
  title        = {Do Emotions Really Affect Argument Convincingness? {A} Dynamic Approach
                  with LLM-based Manipulation Checks},
  journal      = {CoRR},
  volume       = {abs/2503.00024},
  year         = {2025}
}

@inproceedings{DBLP:conf/argmining/PotashFH19,
  author       = {Peter Potash and
                  Adam Ferguson and
                  Timothy J. Hazen},
  title        = {Ranking Passages for Argument Convincingness},
  booktitle    = {ArgMining@ACL},
  pages        = {146--155},
  publisher    = {Association for Computational Linguistics},
  year         = {2019}
}

@inproceedings{DBLP:conf/argmining/GuWXFLH18,
  author       = {Yunfan Gu and
                  Zhongyu Wei and
                  Maoran Xu and
                  Hao Fu and
                  Yang Liu and
                  Xuanjing Huang},
  title        = {Incorporating Topic Aspects for Online Comment Convincingness Evaluation},
  booktitle    = {ArgMining@EMNLP},
  pages        = {97--104},
  publisher    = {Association for Computational Linguistics},
  year         = {2018}
}

@inproceedings{DBLP:conf/eacl/ChalaguineS17,
  author       = {Lisa Andreevna Chalaguine and
                  Claudia Schulz},
  title        = {Assessing Convincingness of Arguments in Online Debates with Limited
                  Number of Features},
  booktitle    = {{EACL} (Student Research Workshop)},
  pages        = {75--83},
  publisher    = {Association for Computational Linguistics},
  year         = {2017}
}

@inproceedings{DBLP:conf/cmna/Palmieri24,
  author       = {Rudi Palmieri},
  title        = {From Loci to Critical Questions: an {AMT} Approach to Argument Evaluation.
                  Insights from the Domain of Corporate Controversies},
  booktitle    = {CMNA@COMMA},
  series       = {{CEUR} Workshop Proceedings},
  volume       = {3769},
  pages        = {81--89},
  publisher    = {CEUR-WS.org},
  year         = {2024}
}

@inproceedings{DBLP:conf/emnlp/Ruiz-DolzHG23,
  author       = {Ramon Ruiz{-}Dolz and
                  Stella Heras and
                  Ana Garc{\'{\i}}a{-}Fornes},
  title        = {Automatic Debate Evaluation with Argumentation Semantics and Natural
                  Language Argument Graph Networks},
  booktitle    = {{EMNLP}},
  pages        = {6030--6040},
  publisher    = {Association for Computational Linguistics},
  year         = {2023}
}

@inproceedings{DBLP:conf/cogsci/Baccini022,
  author       = {Edoardo Baccini and
                  Stephan Hartmann},
  title        = {The Myside Bias in Argument Evaluation: {A} Bayesian Model},
  booktitle    = {CogSci},
  publisher    = {cognitivesciencesociety.org},
  year         = {2022}
}

@inproceedings{DBLP:conf/flairs/FavreauZB22,
  author       = {Charles{-}Olivier Favreau and
                  Amal Zouaq and
                  Sameer Bhatnagar},
  title        = {Learning to Rank with {BERT} for Argument Quality Evaluation},
  booktitle    = {{FLAIRS}},
  publisher    = {Florida Online Journals},
  year         = {2022}
}

@inproceedings{DBLP:conf/chi/DianaSK20,
  author       = {Nicholas Diana and
                  John C. Stamper and
                  Ken Koedinger},
  title        = {Towards Value-Adaptive Instruction: {A} Data-Driven Method for Addressing
                  Bias in Argument Evaluation Tasks},
  booktitle    = {{CHI}},
  pages        = {1--11},
  publisher    = {{ACM}},
  year         = {2020}
}

@inproceedings{DBLP:conf/icls/WinklePN20,
  author       = {Michael Van Winkle and
                  LeAnn Putney and
                  E. Michael Nussbaum},
  title        = {Teachers' Use of Critical Questions for Argument Evaluation},
  booktitle    = {{ICLS}},
  publisher    = {International Society of the Learning Sciences},
  year         = {2020}
}

@article{DBLP:journals/ijar/BorgB24,
  author       = {Annemarie Borg and
                  Floris Bex},
  title        = {Minimality, necessity and sufficiency for argumentation and explanation},
  journal      = {Int. J. Approx. Reason.},
  volume       = {168},
  pages        = {109143},
  year         = {2024}
}

@inproceedings{DBLP:conf/naacl/LiuFC24,
  author       = {Xiao Liu and
                  Yansong Feng and
                  Kai{-}Wei Chang},
  title        = {{CASA:} Causality-driven Argument Sufficiency Assessment},
  booktitle    = {{NAACL-HLT}},
  pages        = {5282--5302},
  publisher    = {Association for Computational Linguistics},
  year         = {2024}
}

@inproceedings{DBLP:conf/argmining/GurckeAW21,
  author       = {Timon Gurcke and
                  Milad Alshomary and
                  Henning Wachsmuth},
  title        = {Assessing the Sufficiency of Arguments through Conclusion Generation},
  booktitle    = {ArgMining@EMNLP},
  pages        = {67--77},
  publisher    = {Association for Computational Linguistics},
  year         = {2021}
}

@article{DBLP:journals/mta/WangPW24,
  author       = {Hei{-}Chia Wang and
                  Cendra Devayana Putra and
                  Chia{-}Ying Wu},
  title        = {A recurrent stick breaking topic model for argument stance detection},
  journal      = {Multim. Tools Appl.},
  volume       = {83},
  number       = {13},
  pages        = {38241--38266},
  year         = {2024}
}

@inproceedings{DBLP:conf/sbp-brims/HosseiniaDM19,
  author       = {Marjan Hosseinia and
                  Eduard C. Dragut and
                  Arjun Mukherjee},
  title        = {Pro/Con: Neural Detection of Stance in Argumentative Opinions},
  booktitle    = {SBP-BRiMS},
  series       = {Lecture Notes in Computer Science},
  volume       = {11549},
  pages        = {21--30},
  publisher    = {Springer},
  year         = {2019}
}

@inproceedings{DBLP:conf/iclr/WagnerBZH25,
  author       = {Stefan Sylvius Wagner and
                  Maike Behrendt and
                  Marc Ziegele and
                  Stefan Harmeling},
  title        = {The Power of LLM-Generated Synthetic Data for Stance Detection in
                  Online Political Discussions},
  booktitle    = {{ICLR}},
  publisher    = {OpenReview.net},
  year         = {2025}
}

@article{DBLP:journals/concurrency/LiYWW24,
  author       = {Hao Li and
                  Wu Yang and
                  Wei Wang and
                  Huanran Wang},
  title        = {Neural network approaches for rumor stance detection: Simulating complex
                  rumor propagation systems},
  journal      = {Concurr. Comput. Pract. Exp.},
  volume       = {36},
  number       = {16},
  year         = {2024}
}

@article{DBLP:journals/ijon/ZhangLZL24,
  author       = {Hao Zhang and
                  Yizhou Li and
                  Tuanfei Zhu and
                  Chuang Li},
  title        = {Commonsense-based adversarial learning framework for zero-shot stance
                  detection},
  journal      = {Neurocomputing},
  volume       = {563},
  pages        = {126943},
  year         = {2024}
}

@article{DBLP:journals/jbi/DavydovaYT24,
  author       = {Vera Davydova and
                  Huabin Yang and
                  Elena Tutubalina},
  title        = {Data and models for stance and premise detection in {COVID-19} tweets:
                  Insights from the Social Media Mining for Health {(SMM4H)} 2022 shared
                  task},
  journal      = {J. Biomed. Informatics},
  volume       = {149},
  pages        = {104555},
  year         = {2024}
}

@article{DBLP:journals/peerj-cs/LeeLK24,
  author       = {Woojin Lee and
                  Jaewook Lee and
                  Harksoo Kim},
  title        = {{LOGIC:} LLM-originated guidance for internal cognitive improvement
                  of small language models in stance detection},
  journal      = {PeerJ Comput. Sci.},
  volume       = {10},
  pages        = {e2585},
  year         = {2024}
}

@article{DBLP:journals/snam/CharfiBAAAZ24,
  author       = {Anis Charfi and
                  Mabrouka Bessghaier and
                  Andria Atalla and
                  Raghda Akasheh and
                  Sara Al{-}Emadi and
                  Wajdi Zaghouani},
  title        = {Stance detection in Arabic with a multi-dialectal cross-domain stance
                  corpus},
  journal      = {Soc. Netw. Anal. Min.},
  volume       = {14},
  number       = {1},
  pages        = {161},
  year         = {2024}
}

@inproceedings{DBLP:conf/acl/ZhaoLCZ24,
  author       = {Chenye Zhao and
                  Yingjie Li and
                  Cornelia Caragea and
                  Yue Zhang},
  title        = {ZeroStance: Leveraging ChatGPT for Open-Domain Stance Detection via
                  Dataset Generation},
  booktitle    = {{ACL} (Findings)},
  pages        = {13390--13405},
  publisher    = {Association for Computational Linguistics},
  year         = {2024}
}

@inproceedings{DBLP:conf/naacl/LiZLGWZLWX25,
  author       = {Ang Li and
                  Jingqian Zhao and
                  Bin Liang and
                  Lin Gui and
                  Hui Wang and
                  Xi Zeng and
                  Xingwei Liang and
                  Kam{-}Fai Wong and
                  Ruifeng Xu},
  title        = {Mitigating Biases of Large Language Models in Stance Detection with
                  Counterfactual Augmented Calibration},
  booktitle    = {{NAACL} (Long Papers)},
  pages        = {7075--7092},
  publisher    = {Association for Computational Linguistics},
  year         = {2025}
}

@article{DBLP:journals/ml/GambiniSFT24,
  author       = {Margherita Gambini and
                  Caterina Senette and
                  Tiziano Fagni and
                  Maurizio Tesconi},
  title        = {Evaluating large language models for user stance detection on {X}
                  (Twitter)},
  journal      = {Mach. Learn.},
  volume       = {113},
  number       = {10},
  pages        = {7243--7266},
  year         = {2024}
}

@inproceedings{DBLP:conf/icic/GuoJL24,
  author       = {Mengzhuo Guo and
                  Xiaorui Jiang and
                  Yong Liao},
  title        = {Improving Zero-Shot Stance Detection by Infusing Knowledge from Large
                  Language Models},
  booktitle    = {{ICIC} {(13)}},
  series       = {Lecture Notes in Computer Science},
  volume       = {14874},
  pages        = {121--132},
  publisher    = {Springer},
  year         = {2024}
}

@inproceedings{DBLP:conf/nlpcc/MaWXZZ24,
  author       = {Junxia Ma and
                  Changjiang Wang and
                  Hanwen Xing and
                  Dongming Zhao and
                  Yazhou Zhang},
  title        = {Chain of Stance: Stance Detection with Large Language Models},
  booktitle    = {{NLPCC} {(5)}},
  series       = {Lecture Notes in Computer Science},
  volume       = {15363},
  pages        = {82--94},
  publisher    = {Springer},
  year         = {2024}
}

@inproceedings{DBLP:conf/wanlp/AlghaslanA24,
  author       = {Mamoun Alghaslan and
                  Khaled Almutairy},
  title        = {{MGKM} at StanceEval2024 Fine-Tuning Large Language Models for Arabic
                  Stance Detection},
  booktitle    = {ArabicNLP},
  pages        = {816--822},
  publisher    = {Association for Computational Linguistics},
  year         = {2024}
}

@inproceedings{DBLP:conf/wanlp/ShuklaVK24,
  author       = {Ishaan Shukla and
                  Ankit Vaidya and
                  Geetanjali Kale},
  title        = {{PICT} at StanceEval2024: Stance Detection in Arabic using Ensemble
                  of Large Language Models},
  booktitle    = {ArabicNLP},
  pages        = {837--841},
  publisher    = {Association for Computational Linguistics},
  year         = {2024}
}

@article{DBLP:journals/corr/abs-2402-14296,
  author       = {Ang Li and
                  Jingqian Zhao and
                  Bin Liang and
                  Lin Gui and
                  Hui Wang and
                  Xi Zeng and
                  Kam{-}Fai Wong and
                  Ruifeng Xu},
  title        = {Mitigating Biases of Large Language Models in Stance Detection with
                  Calibration},
  journal      = {CoRR},
  volume       = {abs/2402.14296},
  year         = {2024}
}

@article{DBLP:journals/corr/abs-2408-04649,
  author       = {Junxia Ma and
                  Changjiang Wang and
                  Hanwen Xing and
                  Dongming Zhao and
                  Yazhou Zhang},
  title        = {Chain of Stance: Stance Detection with Large Language Models},
  journal      = {CoRR},
  volume       = {abs/2408.04649},
  year         = {2024}
}

@article{DBLP:journals/corr/abs-2409-00222,
  author       = {Abu Ubaida Akash and
                  Ahmed Fahmy and
                  Amine Trabelsi},
  title        = {Can Large Language Models Address Open-Target Stance Detection?},
  journal      = {CoRR},
  volume       = {abs/2409.00222},
  year         = {2024}
}

@article{DBLP:journals/corr/abs-2411-14720,
  author       = {Luhang Sun and
                  Varsha Pendyala and
                  Yun{-}Shiuan Chuang and
                  Shanglin Yang and
                  Jonathan Feldman and
                  Andrew Zhao and
                  Munmun De Choudhury and
                  Sijia Yang and
                  Dhavan Shah},
  title        = {Optimizing Social Media Annotation of {HPV} Vaccine Skepticism and
                  Misinformation Using Large Language Models: An Experimental Evaluation
                  of In-Context Learning and Fine-Tuning Stance Detection Across Multiple
                  Models},
  journal      = {CoRR},
  volume       = {abs/2411.14720},
  year         = {2024}
}

@inproceedings{DBLP:conf/acl/0074WSBMZZWHLN24,
  author       = {Hao Li and
                  Yuping Wu and
                  Viktor Schlegel and
                  Riza Batista{-}Navarro and
                  Tharindu Madusanka and
                  Iqra Zahid and
                  Jiayan Zeng and
                  Xiaochi Wang and
                  Xinran He and
                  Yizhi Li and
                  Goran Nenadic},
  title        = {Which Side Are You On? {A} Multi-task Dataset for End-to-End Argument
                  Summarisation and Evaluation},
  booktitle    = {{ACL} (Findings)},
  pages        = {133--150},
  publisher    = {Association for Computational Linguistics},
  year         = {2024}
}

@inproceedings{DBLP:conf/emnlp/IvanovaHN24,
  author       = {Rositsa V. Ivanova and
                  Thomas Huber and
                  Christina Niklaus},
  title        = {Let's discuss! Quality Dimensions and Annotated Datasets for Computational
                  Argument Quality Assessment},
  booktitle    = {{EMNLP}},
  pages        = {20749--20779},
  publisher    = {Association for Computational Linguistics},
  year         = {2024}
}

@inproceedings{DBLP:conf/acl/JoshiPH23,
  author       = {Omkar Joshi and
                  Priya Pitre and
                  Yashodhara Haribhakta},
  title        = {ArgAnalysis35K : {A} large-scale dataset for Argument Quality Analysis},
  booktitle    = {{ACL} {(1)}},
  pages        = {13916--13931},
  publisher    = {Association for Computational Linguistics},
  year         = {2023}
}

@inproceedings{DBLP:conf/acl/RuggeriMG23,
  author       = {Federico Ruggeri and
                  Mohsen Mesgar and
                  Iryna Gurevych},
  title        = {A Dataset of Argumentative Dialogues on Scientific Papers},
  booktitle    = {{ACL} {(1)}},
  pages        = {7684--7699},
  publisher    = {Association for Computational Linguistics},
  year         = {2023}
}

@article{DBLP:journals/corr/abs-2307-02340,
  author       = {Timo Pierre Schrader and
                  Teresa B{\"{u}}rkle and
                  Sophie Henning and
                  Sherry Tan and
                  Matteo Finco and
                  Stefan Gr{\"{u}}newald and
                  Maira Indrikova and
                  Felix Hildebrand and
                  Annemarie Friedrich},
  title        = {MuLMS-AZ: An Argumentative Zoning Dataset for the Materials Science
                  Domain},
  journal      = {CoRR},
  volume       = {abs/2307.02340},
  year         = {2023}
}

@inproceedings{DBLP:conf/acl/ChengBHYZS22,
  author       = {Liying Cheng and
                  Lidong Bing and
                  Ruidan He and
                  Qian Yu and
                  Yan Zhang and
                  Luo Si},
  title        = {{IAM:} {A} Comprehensive and Large-Scale Dataset for Integrated Argument
                  Mining Tasks},
  booktitle    = {{ACL} {(1)}},
  pages        = {2277--2287},
  publisher    = {Association for Computational Linguistics},
  year         = {2022}
}

@article{DBLP:journals/corr/abs-2205-11472,
  author       = {Benjamin Schiller and
                  Johannes Daxenberger and
                  Iryna Gurevych},
  title        = {On the Effect of Sample and Topic Sizes for Argument Mining Datasets},
  journal      = {CoRR},
  volume       = {abs/2205.11472},
  year         = {2022}
}

@inproceedings{DBLP:conf/aaai/GretzFCTLAS20,
  author       = {Shai Gretz and
                  Roni Friedman and
                  Edo Cohen{-}Karlik and
                  Assaf Toledo and
                  Dan Lahav and
                  Ranit Aharonov and
                  Noam Slonim},
  title        = {A Large-Scale Dataset for Argument Quality Ranking: Construction and
                  Analysis},
  booktitle    = {{AAAI}},
  pages        = {7805--7813},
  publisher    = {{AAAI} Press},
  year         = {2020}
}

@inproceedings{DBLP:conf/emnlp/Toledo-RonenOBS20,
  author       = {Orith Toledo{-}Ronen and
                  Matan Orbach and
                  Yonatan Bilu and
                  Artem Spector and
                  Noam Slonim},
  title        = {Multilingual Argument Mining: Datasets and Analysis},
  booktitle    = {{EMNLP} (Findings)},
  series       = {Findings of {ACL}},
  volume       = {{EMNLP} 2020},
  pages        = {303--317},
  publisher    = {Association for Computational Linguistics},
  year         = {2020}
}

@inproceedings{DBLP:conf/emnlp/ToledoGCFVLJAS19,
  author       = {Assaf Toledo and
                  Shai Gretz and
                  Edo Cohen{-}Karlik and
                  Roni Friedman and
                  Elad Venezian and
                  Dan Lahav and
                  Michal Jacovi and
                  Ranit Aharonov and
                  Noam Slonim},
  title        = {Automatic Argument Quality Assessment - New Datasets and Methods},
  booktitle    = {{EMNLP/IJCNLP} {(1)}},
  pages        = {5624--5634},
  publisher    = {Association for Computational Linguistics},
  year         = {2019}
}
